\documentclass[journal]{IEEEtai}

\usepackage[colorlinks,urlcolor=blue,linkcolor=blue,citecolor=blue]{hyperref}

\usepackage{color,array}
\usepackage{amsmath, amssymb}
\usepackage{graphicx}
\usepackage{booktabs}
\usepackage{algorithm, algorithmicx, algpseudocode}
\usepackage{comment}
\usepackage{bm}
\usepackage{multirow}
\usepackage{subcaption} 

\newtheorem{remark}{Remark}
\newtheorem{corollary}{Corollary}
\newtheorem{proposition}{Proposition}
\newcommand{\autoformer}{\texttt{Autoformer}}
\newcommand{\informer}{\texttt{Informer}}
\newcommand{\patchtst}{\texttt{PatchTST}}

\newcommand{\fedformer}{\texttt{FEDformer}}
\newcommand{\pyraformer}{\texttt{Pyraformer}}

\newcommand{\dkoopformer}{\texttt{DeepKoopFormer}}

\newcommand{\sigmoid}{\sigma}

\usepackage{hyperref}

\begin{document}

\title{\dkoopformer: A Koopman Enhanced Transformer Based Architecture for Time Series Forecasting}

\author{Ali Forootani, \IEEEmembership{Senior, IEEE}, Mohammad Khosravi, \IEEEmembership{Member, IEEE}, Masoud Barati
\thanks{Ali Forootani is with Helmholtz Center for Environmental Research-UFZ, Permoserstraße 15, 04318 Leipzig, Germany, \texttt{Email: aliforootani@ieee.org/ali.forootani@ufz.de}.}
\thanks{
Mohammad Khosravi is with Delft Center for Systems and Control, Mekelweg 2, Delft, 2628 CD, The Netherlands, \texttt{Email: mohammad.khosravi@tudelft.nl}
}
\thanks{
	Masoud Barati is with School of Engineering, University of Pittsburgh Swanson, O'Hara Street Benedum Hall of Engineering, Pittsburgh, 3700, Pennsylvania, United States
	\texttt{Email: masoud.barati@pitt.edu}
	}
}


\maketitle
\begin{abstract}
Time series forecasting plays a vital role across scientific, industrial, and environmental domains, especially when dealing with high-dimensional and nonlinear systems. While Transformer-based models have recently achieved state-of-the-art performance in long-range forecasting, they often suffer from interpretability issues and instability in the presence of noise or dynamical uncertainty. In this work, we propose \dkoopformer, a principled forecasting framework that combines the representational power of Transformers with the theoretical rigor of Koopman operator theory. Our model features a modular encoder--propagator--decoder structure, where temporal dynamics are learned via a spectrally constrained, linear Koopman operator in a latent space. We impose structural guarantees---such as bounded spectral radius, Lyapunov-based energy regularization, and orthogonal parameterization---to ensure stability and interpretability. Comprehensive evaluations are conducted on both synthetic dynamical systems, real-world climate dataset (wind speed and surface pressure), financial time series (cryptocurrency), and electricity generation dataset using the \texttt{Python} package that is prepared for this purpose. Across all experiments, \dkoopformer~consistently outperforms standard LSTM and baseline Transformer models in terms of accuracy, robustness to noise, and long-term forecasting stability. These results establish \dkoopformer~as a flexible, interpretable, and robust framework for forecasting in high-dimensional and dynamical settings.

\end{abstract}



\section{Introduction}
\label{sec:intro}

Time series forecasting is a foundational task in science and engineering, with critical applications ranging from retail demand prediction \cite{Croston1972,wen2017multi,salinas2020deepar}, traffic flow management \cite{lv2014traffic,li2018diffusion}, and energy balancing \cite{DIMOULKAS2019,Saxena2019,forootani2024climate} to financial volatility modeling \cite{callot2017modeling,neurips_18_long_tail}. The increasing availability of high-dimensional, high-frequency, and often non-stationary data has rendered traditional statistical methods insufficient for capturing the complex dependencies and dynamics of modern time series \cite{hyndman2018forecasting}.

Deep learning models have revolutionized this landscape by offering scalable and expressive alternatives. Early architectures such as Recurrent Neural Networks (RNNs) \cite{hochreiter1997long,salinas2020deepar} and Convolutional Neural Networks (CNNs) \cite{lecun1995convolutional,BaiTCN2018} laid the foundation by automating temporal feature extraction. More recently, Transformer-based models \cite{wu2021autoformer,haoyietal-informer-2021,nie2022time} have emerged as the dominant paradigm due to their ability to model long-range dependencies via self-attention. These models have demonstrated state-of-the-art performance in large-scale competitions such as M4 and M5 \cite{smyl2018m4,makridakis2021m5}.

To address specific forecasting challenges, a range of specialized Transformer architectures have been proposed. \informer \cite{informer} introduces \texttt{ProbSparse} attention to focus computation on the most informative queries and reduce redundancy in the attention matrix. \autoformer~\cite{autoformer} incorporates a decomposition block that explicitly separates trend and seasonal components, while \fedformer \cite{fedformer} replaces time-domain attention with Fourier-based decomposition to improve scalability. \pyraformer~\cite{pyraformer} employs a pyramidal attention structure to capture hierarchical temporal dependencies. Although these architectures differ in design, they share a common objective: achieving high predictive accuracy while controlling computational complexity.

However, recent work has questioned the necessity of such architectural complexity. In \cite{dlinear}, the authors propose \texttt{DLinear}, a simple channel-wise linear model that surprisingly outperforms many deep Transformer variants on several benchmark datasets. This finding suggests that much of the predictive power in time series may stem from capturing channel-specific dynamics rather than from complex attention-based mechanisms. In response to this, \patchtst~\cite{nie2022time} proposes a streamlined Transformer architecture that improves generalization by (i) employing \emph{patching} to capture local and semantic dependencies, and (ii) adopting a \emph{channel-independent} learning scheme, where each variable is processed separately to reduce overfitting and preserve signal-specific structure \cite{multichannel,dlinear}.

Despite these innovations, several persistent challenges remain. Deep models often lack interpretability, exhibit sensitivity to distributional shifts \cite{kim2022reversible,kuznetsov2020discrepancy,Liu2022NonstationaryTR}, and typically do not incorporate prior knowledge about the underlying physical or dynamical systems. In many real-world applications, robustness and interpretability are as important as raw accuracy, especially when forecasts are used in high-stakes decision-making.

To address these limitations, recent trends have sought to embed physical priors and system dynamics into learning architectures. Physics-inspired models such as Neural ODEs \cite{chen19,vialard20} and hybrid deep-statistical approaches \cite{wang2019deepfactors,sen2019think,smyl2018m4} attempt to bridge the gap between data-driven learning and model-based reasoning. Yet these methods often assume fixed dynamics and face scalability issues, especially in high-dimensional or multivariate settings.

This motivates the development of forecasting frameworks that not only leverage the expressive power of deep learning but also impose structure rooted in dynamical systems theory.


\subsection{Contributions}

We introduce \dkoopformer, a principled and modular neural forecasting framework that integrates Transformer-based sequence modeling with Koopman operator theory. Compared with prior Transformer forecasters and Koopman‐based models, our framework delivers stronger theoretical guarantees and greater practical flexibility. Our main contributions are:

\noindent (i) \text{Koopman-Enhanced Transformer Architecture.}  
    We introduce an encoder–propagator–decoder design in which a Transformer encoder supplies latent features that evolve linearly under a learned Koopman operator before being linearly decoded to the next observation. This modular decomposition cleanly separates representation learning from temporal evolution, improving interpretability and stability \cite{koopman1931hamiltonian,Mezic2005nd}.
    
\noindent (ii) \text{Spectrally Constrained Koopman Propagator.}  
    The latent transition matrix is parameterised by an orthogonal–diagonal–orthogonal (ODO) factorisation whose diagonal entries are strictly confined inside the unit circle. This enforces normality, bounds gradients, and guarantees exponential decay of latent trajectories, distinguishing \dkoopformer~from unconstrained Koopman approximations \cite{lusch2018deep,takeishi2017learning}.

\noindent (iii) \text{Provable Forecast Robustness.}  
    Leveraging the spectral bound, we derive a closed-form, geometrically decaying upper bound on error growth under input perturbations. The result yields the first formal forecast-error guarantee for a Transformer-based time-series model.

\noindent (iv) \text{Energy-Regularised Latent Stability.}  
    A Lyapunov-inspired penalty discourages transient energy increases in the latent space, complementing the spectral constraint and further stabilizing both training and short-horizon predictions.

\noindent (v) \text{Capacity–Stability Decoupling.}  
    Representation capacity (latent width) and dynamical stability (spectral radius) are controlled by two orthogonal hyper-parameters, greatly simplifying hyper-parameter search across diverse forecasting regimes.

\noindent (vi) \text{Numerical Robustness via Orthogonality Projection.}  
    Exact Householder–QR re-orthogonalisation of Koopman factors on every forward pass preserves unit-norm bases, prevents gradient blow-up, and maintains well-conditioned optimisation.

\noindent (vii) \text{Predictive-Only Training Objective.}  
    In contrast to Koopman autoencoders that balance reconstruction and prediction losses \cite{lusch2018deep}, \dkoopformer~ optimises a pure forecasting loss, eliminating invertibility requirements and allowing more expressive encoders.

\noindent (viii) \text{Channel-Independent Encoding.}  
    Following the \emph{channel-independent} strategy of \patchtst~\cite{nie2022time}, each variable is embedded separately, reducing overfitting risk in high-dimensional multivariate settings \cite{dlinear,multichannel}.

\noindent (ix) \text{Unified Koopman Benchmark Suite.}  
    We implement Koopman-enhanced variants of \patchtst~\cite{nie2022time}, \autoformer \cite{wu2021autoformer}, and \informer~\cite{haoyietal-informer-2021}, releasing an High Performance Computing (HPC) in a \texttt{Python} package\footnote{\url{https://github.com/Ali-Forootani/deepkoopformer}}-ready benchmark with complete training/testing scripts and ablation studies. To evaluate the effectiveness and robustness of the proposed unified framework, we conduct extensive experiments on a diverse suite of datasets, encompassing both synthetic and real-world scenarios. Synthetic benchmarks included the nonlinear Van der Pol and Lorenz systems, each augmented with Gaussian noise to simulate realistic dynamics. 

    For real-world evaluation, we utilize high-dimensional datasets from multiple domains, including: the CMIP6 climate projections~\footnote{\url{https://cds.climate.copernicus.eu/datasets/projections-cordex-domains-single-levels?tab=overview}} and ERA5 reanalysis data~\footnote{\url{https://cds.climate.copernicus.eu/datasets}}, focusing on wind speed and surface pressure forecasting over Germany; a financial time series dataset~\footnote{\url{https://github.com/Chisomnwa/Cryptocurrency-Data-Analysis}} for cryptocurrency market analysis; and an electricity generation dataset~\footnote{\url{https://github.com/afshinfaramarzi/Energy-Demand-electricity-price-Forecasting/tree/main}} for modeling energy supply dynamics.

    These datasets collectively span chaotic, periodic, and stochastic regimes, allowing for a comprehensive assessment of model accuracy, stability, and generalization across domains.

Compared to existing Koopman-based forecasting models \cite{lusch2018deep,takeishi2017learning,azencot2020forecasting}, \dkoopformer~is the first to offer provable spectral decay, Lyapunov-stable propagation, normality of the transition matrix, and closed-form error control—all within a general, extensible encoder–propagator–decoder pipeline aligned with the goals of physics-informed forecasting \cite{PIML2021,Truong2023} and interpretable dynamical systems modeling \cite{Brunton2016plosone,Korda2016arxiv}.

This paper is organized as follows. Section~\ref{preliminaries_sec} provides the preliminaries required for time series forecasting. In Section~\ref{sec:transformer_variants}, we present extended variants of the \patchtst, \autoformer, and \informer~architectures for multivariate time series forecasting. Section~\ref{koopman} is dedicated to the \dkoopformer~building-block architecture and its associated analysis. Numerical simulations are presented in Section~\ref{sec:simulations} to evaluate the performance of \dkoopformer. Finally, we conclude the article in Section~\ref{conclusion_sec}.


\section{Preliminaries and Notation}
\label{preliminaries_sec}

We begin by formalizing the multivariate time series forecasting task and defining key components used in recent Transformer-based forecasting models.

\paragraph{Multivariate Time Series}
Let 
\[
\bm x = [\bm x_1, \bm x_2, \dots, \bm x_T] \in \mathbb{R}^{T \times d}
\]
denote a multivariate time series with $d$ observed variables and horizon length $T$, where each observation $\bm x_t \in \mathbb{R}^{d}$ corresponds to measurements at time step $t$.

\paragraph{Forecasting Setup}
Given a historical context of length $P$, the objective is to predict the next $H$ future steps. For any index $t \in \{1, \dots, T - P - H + 1\}$, we define:
\begin{align}
\bm X_t &= [\bm x_t, \bm x_{t+1}, \dots, \bm x_{t+P-1}] \in \mathbb{R}^{P \times d}, \\
\bm Y_t &= [\bm x_{t+P}, \bm x_{t+P+1}, \dots, \bm x_{t+P+H-1}] \in \mathbb{R}^{H \times d}.
\end{align}
The goal is to learn a mapping \( f_\theta: \mathbb{R}^{P \times d} \to \mathbb{R}^{H \times d} \) such that
\begin{equation}
\hat{\bm Y}_t = f_\theta(\bm X_t),
\end{equation}
where \(\hat{\bm Y}_t\) denotes the predicted sequence.

\paragraph{Input Embedding}
To project raw inputs to a model-specific latent space of dimension $d_{\text{model}}$, we apply a channel-wise linear embedding:
\[
\bm Z_t = \bm X_t \cdot \bm W_e, \qquad \bm W_e \in \mathbb{R}^{d \times d_{\text{model}}}, \quad \bm Z_t \in \mathbb{R}^{P \times d_{\text{model}}}.
\]

\paragraph{Positional Encoding}
Since Transformers lack an inherent notion of temporal order, positional encodings $\text{PE} \in \mathbb{R}^{P \times d_{\text{model}}}$ are added to the input embeddings:
\[
\bm Z_t^{\text{pos}} = \bm Z_t + \text{PE},
\]
where $\text{PE}$ may be either:
\begin{itemize}
    \item \textit{Fixed (sinusoidal)}~\cite{vaswani2017attention}:
    \begin{align*}
        \text{PE}_{(p,2k)} &= \sin\left(\frac{p}{10000^{2k/d_{\text{model}}}}\right),\\
    \text{PE}_{(p,2k+1)} &= \cos\left(\frac{p}{10000^{2k/d_{\text{model}}}}\right),
    \end{align*}

    \item \textit{Learnable}, optimized during training.
\end{itemize}

\paragraph{Transformer Encoder}
The encoder processes the position-aware sequence \(\bm Z_t^{\text{pos}} \in \mathbb{R}^{P \times d_{\text{model}}}\) through multiple layers of multi-head self-attention (MSA) and feedforward networks (FFN). A typical layer outputs:
\begin{equation}
\bm H^{(e)} = \text{LayerNorm}\left(\bm Z + \text{FFN}\left(\text{MSA}(\bm Z)\right)\right), \quad \bm H^{(e)} \in \mathbb{R}^{P \times d_{\text{model}}}.
\end{equation}

\paragraph{Pooling and Aggregation}
To extract a fixed-length feature from the sequence, pooling operations are applied:
\begin{equation}
\bm h^{(e)} = \text{AvgPool}(\bm H^{(e)}) = \frac{1}{P} \sum_{p=1}^{P} \bm H^{(e)}_p \in \mathbb{R}^{d_{\text{model}}}.
\end{equation}

\paragraph{Output Projection}
The final forecast is obtained by projecting the latent state to the output horizon:
\[
\hat{\bm Y}_t = \bm h^{(e)} \cdot \bm W_o, \qquad \bm W_o \in \mathbb{R}^{d_{\text{model}} \times (H \cdot d)}.
\]
The output is reshaped to \(\hat{\bm Y}_t \in \mathbb{R}^{H \times d}\).

Several Transformer-based models extend this general framework such as:

(i) \informer~\cite{informer}: Replaces full self-attention with \textit{ProbSparse attention}, selecting a subset of key–value pairs per query to reduce complexity from \(O(P^2)\) to \(O(P \log P)\), with attention sparsity scored via:
    \[
    M(q_i) = \max_j \alpha_{ij} - \text{mean}_j \alpha_{ij}, \quad \alpha_{ij} = \frac{q_i^\top k_j}{\sqrt{d_{\text{model}}}};\]

(ii) \autoformer~\cite{autoformer}: Decomposes the input into trend and seasonal components using a moving average (MA) filter of kernel size $k$:
\[
\bm X_t = \bm X_t^{\text{trend}} + \bm X_t^{\text{seasonal}}, \quad \bm X_t^{\text{trend}} = \text{MA}_k(\bm X_t);
\]
The model separately encodes seasonal and trend dynamics with residual learning.
    
(iii) \patchtst~\cite{nie2022time}: Treats time series as a sequence of non-overlapping patches, enabling local pattern extraction akin to Vision Transformers. The input is reshaped into $N$ patches:
    \[
    \bm X_t^{\text{patch}} \in \mathbb{R}^{N \times (p \cdot d)}, \quad \text{with } p = P/N,
    \]
    and processed via a 1D Transformer for patch-wise temporal modeling.    

We denote by $d_{\mathrm{model}}$ the hidden width of the Transformer encoder, which determines the dimensionality of the internal representation used for attention and feedforward operations.

Moreover, to enable efficient training and inference, we organize training samples into batches of size $B$. Each batch contains a set of input–output pairs $(\bm X_t, \bm Y_t)$, where:
\[
\bm X \in \mathbb{R}^{B \times P \times d}, \qquad \bm Y \in \mathbb{R}^{B \times H \times d}.
\]
This notation is used consistently throughout our algorithmic descriptions.


\section{Transformer Variants for Time-Series Forecasting: Architecture and Computational Complexity}
\label{sec:transformer_variants}
Building on our earlier univariate benchmark study \cite{forootanicompactformer2025}, we now \emph{extend and unify} three leading Transformer forecasting frameworks—\patchtst, \informer, and \autoformer—for \emph{multivariate} time–series prediction.  For every framework we introduce a coherent trio of architectural variants (\textit{Minimal}, \textit{Standard}, and \textit{Full}) and analyse them through a single analytical lens that juxtaposes (i) their cross-channel interaction strategy (channel-independent patching, ProbSparse attention, or trend–seasonal decomposition), (ii) representational capacity, and (iii) asymptotic complexity as a function of both sequence length~$P$ and variable dimension~$d$.  This taxonomy illuminates how design choices—such as patch tokenisation, sparse attention, and decomposition blocks—affect accuracy, robustness, and computational efficiency in multivariate settings, providing a transparent guide for practitioners choosing among competing Transformer architectures.

\subsection{Minimal Variants: Lightweight Baselines}
Minimal variants serve as baseline encoder-only models emphasizing simplicity and computational efficiency.

\paragraph{\patchtst~Minimal} Inputs $\bm X_t \in \mathbb{R}^{P \times d}$ are divided into $N$ non-overlapping patches:
\begin{equation*}
\bm X_t^{\mathrm{patch}} \in \mathbb{R}^{N \times (p \cdot d)}, \quad \text{with } p = P / N,
\end{equation*}
Each patch is projected:
\begin{equation*}
\bm Z_t = \bm X_t^{\mathrm{patch}} \bm W_e \in \mathbb{R}^{N \times d_{\mathrm{model}}},
\end{equation*}
with sinusoidal positional encoding:
\begin{equation*}
\bm Z_t^{\mathrm{pos}} = \bm Z_t + \text{PE} \in \mathbb{R}^{N \times d_{\mathrm{model}}}.
\end{equation*}
The forecast is obtained via Transformer encoding, pooling, and linear projection:
\begin{equation*}
\hat{\bm Y}_t = \text{Reshape}\left(\bm h^{(e)} \bm W_o\right) \in \mathbb{R}^{H \times d},
\end{equation*}
where $\bm h^{(e)} = \text{AvgPool}(\text{TransformerEncoder}(\bm Z_t^{\mathrm{pos}})) \in \mathbb{R}^{d_{\mathrm{model}}}$.

\paragraph{\informer~Minimal} Operates directly on $\bm X_t \in \mathbb{R}^{P \times d}$ with full self-attention:
\begin{align*}
\bm Z_t &= \bm X_t \bm W_e + \text{PE}, \\
\bm h^{(e)} &= \text{AvgPool}(\text{TransformerEncoder}(\bm Z_t)), \\
\hat{\bm Y}_t &= \text{Reshape}\left(\bm h^{(e)} \bm W_o\right) \in \mathbb{R}^{H \times d}.
\end{align*}

\paragraph{\autoformer~Minimal} Applies trend-seasonal decomposition to $\bm X_t$:
\begin{align*}
\bm X_t^{\mathrm{trend}} &= \mathrm{MA}_k(\bm X_t), \\
\bm X_t^{\mathrm{seasonal}} &= \bm X_t - \bm X_t^{\mathrm{trend}},
\end{align*}
and then:
\begin{align*}
\bm Z_t^{\mathrm{seasonal}} &= \bm X_t^{\mathrm{seasonal}} \bm W_e + \text{PE}, \\
\bm h^{(e)} &= \text{AvgPool}(\text{TransformerEncoder}(\bm Z_t^{\mathrm{seasonal}})), \\
\hat{\bm Y}_t &= \text{Reshape}\left(\bm h^{(e)} \bm W_o\right) + \bm X_t^{\mathrm{trend}} \bm W_t \in \mathbb{R}^{H \times d}.
\end{align*}

\paragraph{Complexity (per batch)}
For clarity, we separate the three Minimal variants and make the role of $d$ (input channels) and $d_{\mathrm{model}}$ (latent/projection dimension) explicit:

\begin{align*}
\text{\patchtst: } & \mathcal{O}(N^2 d_{\mathrm{model}} + N d\, d_{\mathrm{model}} + N d_{\mathrm{model}}^2),\\ &N = P / p; \\
\text{\informer: } & \mathcal{O}(P^2 d_{\mathrm{model}} + P d\, d_{\mathrm{model}} + P d_{\mathrm{model}}^2); \\
\text{\autoformer: } & \mathcal{O}(P^2 d_{\mathrm{model}} + P d\, d_{\mathrm{model}} + P d_{\mathrm{model}}^2)\\ &(\mathrm{MA}_k : \mathcal{O}(P d)), \\
\end{align*}
the above complexity formulas for Minimal variants of \patchtst, \informer, and \autoformer~are valid for encoder-only architectures with full self-attention.

\subsection{Standard Variants: Adaptive Encoding and Sparse Attention}
Standard variants introduce learnable positional encodings or efficient sparse attention.

\paragraph{\patchtst~Standard} Replaces fixed positional encoding with a trainable $\text{PE}_{\text{learn}}$:
\begin{equation*}
\bm Z_t^{\mathrm{pos}} = \bm Z_t + \text{PE}_{\text{learn}},
\end{equation*}
all other steps as in Minimal.

\paragraph{\informer~Standard} Uses ProbSparse attention in place of full self-attention:
\begin{align*}
\alpha_{ij} &= \frac{\bm q_i^\top \bm k_j}{\sqrt{d_{\mathrm{model}}}}, \\
M(\bm q_i) &= \max_j \alpha_{ij} - \mathrm{mean}_j \alpha_{ij}, \\
&\text{(select top-$u$ queries with $u = \mathcal{O}(\log P)$)}.
\end{align*}

\paragraph{\autoformer~Standard} Retains encoder-only with improved residual blocks and loss strategies.

\paragraph{Complexity (per batch, Standard Variants)}

\begin{align*}
\text{\patchtst: } & \mathcal{O}(N^2 d_{\mathrm{model}} + N d\, d_{\mathrm{model}} + N d_{\mathrm{model}}^2),\\  &N = P / p; \\
\text{\informer: } & \mathcal{O}(P \log P\, d_{\mathrm{model}} + P d\, d_{\mathrm{model}} + P d_{\mathrm{model}}^2); \\
\text{\autoformer: } & \mathcal{O}(P^2 d_{\mathrm{model}} + P d\, d_{\mathrm{model}} + P d_{\mathrm{model}}^2), \\
\end{align*}
Complexity given for $L=1$ encoder layer; multiply by $L$ for multi-layer models.

\subsection{Full Variants: Sequence-to-Sequence Forecasting}
Full variants use encoder-decoder architectures with cross-attention for richer temporal modeling.

\paragraph{\patchtst~Full}
\begin{align*}
\bm H^{(e)} &= \text{TransformerEncoder}(\bm Z_t^{(e)}), \\
\bm Q^{(d)} &= \bm X_t^{\mathrm{rep}} \bm W_e^{(d)} + \text{PE}^{(d)}, \\
\bm H^{(d)} &= \text{TransformerDecoder}(\bm Q^{(d)}, \bm H^{(e)}), \\
\hat{\bm Y}_t &= \text{Reshape}\left(\bm H^{(d)} \bm W_o\right) \in \mathbb{R}^{H \times d},
\end{align*}
where $\bm X_t^{\mathrm{rep}}$ is the repeated last context value for the decoder input.

\paragraph{\informer~Full} ProbSparse attention in both encoder and decoder:
\begin{align*}
\bm H^{(e)} &= \text{ProbSparseEncoder}(\bm Z_t^{(e)}), \\
\bm H^{(d)} &= \text{ProbSparseDecoder}(\bm Q^{(d)}, \bm H^{(e)}), \\
\hat{\bm Y}_t &= \text{Reshape}\left(\bm H^{(d)} \bm W_o\right).
\end{align*}

\paragraph{\autoformer~Full} Trend-seasonal decomposition and encoder-decoder structure:
\begin{align*}
\bm H^{(e)} &= \text{TransformerEncoder}(\bm Z_t^{\mathrm{seasonal}}), \\
\bm H^{(d)} &= \text{TransformerDecoder}(\bm Q^{(d)}, \bm H^{(e)}), \\
\hat{\bm Y}_t &= \text{Reshape}\left(\bm H^{(d)} \bm W_o\right) + \bm X_t^{\mathrm{trend}} \bm W_t.
\end{align*}

\paragraph{Complexity (Full, per batch)}

\begin{align*}
\text{\patchtst: }&  \\
&\mathcal{O}( N^{2} d_{\mathrm{model}}
\;+\; H^{2} d_{\mathrm{model}}
 \;+\; H N \, d_{\mathrm{model}}
\\ & \;+\; (N+H)\,d_{\mathrm{model}}^{2}), N = P/p \\
\text{\informer: }& \\
&\mathcal{O}(
P \log P\, d_{\mathrm{model}}
\;+\;H \log H\, d_{\mathrm{model}}
 \\ & \;+\;  H P\, d_{\mathrm{model}}
\;+\; (P+H)\,d_{\mathrm{model}}^{2}), \\
\text{\autoformer: }& \\
&\mathcal{O}(
P^{2} d_{\mathrm{model}}
\;+\; H^{2} d_{\mathrm{model}}
\\ & \;+\; H P\, d_{\mathrm{model}}
\;+\; (P+H)\,d_{\mathrm{model}}^{2}),
\end{align*}

Complexity expressions are for a single encoder and decoder layer ($L=1$); for $L$ layers, we should multiply the corresponding encoder/decoder terms by $L$.


This unified taxonomy provides a flexible framework for understanding the modular design of Transformer-based forecasters (see Table \ref{tab:unified_transformer_variants}). Moreover, encoder and decoder procedure of the unified transformer architectures are shown in the Algorithms \ref{alg:unified_encoder}, and \ref{alg:unified_decoder}, respectively in Appendix.

\begin{table*}[!t]
\renewcommand{\arraystretch}{1.3}
\caption{Unified comparison of Transformer variants. $P$: input sequence length; $H$: forecast horizon; $d$: input channel dimension; $d_{\text{model}}$: model width. All complexities are per encoder/decoder layer.}
\label{tab:unified_transformer_variants}
\centering
\begin{tabular}{|c|c|c|c|c|}
\hline
\textbf{Model Family} & \textbf{Variant} & \textbf{Architecture} & \textbf{Key Feature} & \textbf{Time Complexity}   \\
\hline
\multirow{3}{*}{\patchtst}
  & Minimal  & Encoder‐only & Sinusoidal PE & $\mathcal{O}(N^2 d_{\text{model}} + N d d_{\text{model}} + N d_{\text{model}}^2)$  \\
  & Standard & Encoder‐only & Trainable PE & $\mathcal{O}(N^2 d_{\text{model}} + N d d_{\text{model}} + N d_{\text{model}}^2)$  \\
  & Full & Encoder–Decoder & Cross-attn decoder & $\mathcal{O}(N^2 d_{\text{model}} + H^2 d_{\text{model}} + HN d_{\text{model}} + (N + H) d_{\text{model}}^2)$  \\
\hline
\multirow{3}{*}{\informer}
  & Minimal  & Encoder‐only & Full attention baseline       & $\mathcal{O}(P^2 d_{\text{model}} + P d d_{\text{model}} + P d_{\text{model}}^2)$   \\
  & Standard & Encoder‐only & ProbSparse (log‐scale)        & $\mathcal{O}(P \log P\, d_{\text{model}} + P d d_{\text{model}} + P d_{\text{model}}^2)$   \\
  & Full     & Encoder–Decoder      & ProbSparse dual attention & $\mathcal{O}(P \log P\, d_{\text{model}} + H \log H\, d_{\text{model}} + HP d_{\text{model}} + (P + H) d_{\text{model}}^2)$   \\
\hline
\multirow{3}{*}{\autoformer}
  & Minimal  & Encoder‐only & Trend–seasonal (shallow)  & $\mathcal{O}(P^2 d_{\text{model}} + P d d_{\text{model}} + P d_{\text{model}}^2)$   \\
  & Standard & Encoder‐only & Robust encoder w/ MA   & $\mathcal{O}(P^2 d_{\text{model}} + P d d_{\text{model}} + P d_{\text{model}}^2)$   \\
  & Full  & Enc–Dec      & Seq2Seq moving‐average        & $\mathcal{O}(P^2 d_{\text{model}} + H^2 d_{\text{model}} + HP d_{\text{model}} + (P + H) d_{\text{model}}^2)$  \\
\hline
\end{tabular}
\vspace{1mm}

\raggedright
\textbf{Note:} For \patchtst, $N = P/p$ is the number of patches for patch length $p$. All complexities are per encoder/decoder layer.
\end{table*}

\begin{algorithm}[!t]
\caption{Unified Encoder Procedure (\patchtst, \informer, \autoformer)}
\label{alg:unified_encoder}
\begin{algorithmic}[1]
\State \textbf{Input:} Input $\bm X_t \in \mathbb{R}^{B \times P \times d}$; model type $\mathcal{M}$
\State \textbf{Output:} Encoder output $\bm H^{(e)} \in \mathbb{R}^{B \times L \times d_{\text{model}}}$; trend component $\bm X_t^{\text{trend}}$
\If{$\mathcal{M} = \autoformer$}
    \State $\bm X_t^{\text{trend}} \gets \mathrm{MA}_k(\bm X_t)$
    \State $\bm X_t^{\text{seasonal}} \gets \bm X_t - \bm X_t^{\text{trend}}$
    \State $\bm X_{\text{enc}} \gets \bm X_t^{\text{seasonal}}$
\Else
    \State $\bm X_{\text{enc}} \gets \bm X_t$
    \State $\bm X_t^{\text{trend}} \gets \mathbf{0}$
\EndIf
\If{$\mathcal{M} = \patchtst$}
    \State $\bm X_t^{\text{patch}} \gets \mathrm{ReshapePatches}(\bm X_{\text{enc}})$ \Comment{$\mathbb{R}^{B \times N \times (p \cdot d)}$}
    \State $\bm Z^{(e)} \gets \bm X_t^{\text{patch}} \cdot \bm W_e^{(e)} + \text{PE}^{(e)}$
\Else
    \State $\bm Z^{(e)} \gets \bm X_{\text{enc}} \cdot \bm W_e^{(e)} + \text{PE}^{(e)}$
\EndIf
\For{each encoder layer}
    \State Compute $Q, K, V$ from $\bm Z^{(e)}$
    \If{$\mathcal{M} = \informer$}
        \State Select top-$u$ queries via ProbSparse attention
    \Else
        \State Apply full attention: $\mathrm{Softmax}(QK^\top / \sqrt{d_{\text{model}}}) \cdot V$
    \EndIf
    \State Apply FFN, residual connection, and normalization
\EndFor
\State $\bm H^{(e)} \gets$ output of final encoder layer
\State \Return $\bm H^{(e)}, \bm X_t^{\text{trend}}$
\end{algorithmic}
\end{algorithm}

\begin{algorithm}[!t]
\caption{Unified Decoder and Forecasting Procedure (Full Variants Only)}
\label{alg:unified_decoder}
\begin{algorithmic}[1]
\State \textbf{Input:} Encoder output $\bm H^{(e)}$; trend $\bm X_t^{\text{trend}}$; input $\bm X_t \in \mathbb{R}^{B \times P \times d}$; model $\mathcal{M}$; horizon $H$
\State \textbf{Output:} Forecast $\hat{\bm Y}_t \in \mathbb{R}^{B \times H \times d}$
\If{$\mathcal{M} = \autoformer$}
    \State $\bm X_{\text{dec}} \gets \mathbf{0} \in \mathbb{R}^{B \times H \times d}$
\Else
    \State Repeat last step: $\bm X_{\text{dec}} \gets \mathrm{Repeat}(\bm X_{t, -1}, H)$
\EndIf
\State Embed decoder input: $\bm Z^{(d)} \gets \bm X_{\text{dec}} \cdot \bm W_e^{(d)} + \text{PE}^{(d)}$
\For{each decoder layer}
    \State Apply self-attention on $\bm Z^{(d)}$
    \State Apply cross-attention: $\mathrm{Softmax}(QK^\top / \sqrt{d_{\text{model}}}) \cdot V$ using $\bm H^{(e)}$
    \State Apply FFN, residual connection, and normalization
\EndFor
\If{$\mathcal{M} = \autoformer$}
    \State $\hat{\bm Y}_t^{\text{seasonal}} \gets \bm Z^{(d)} \cdot \bm W_o$
    \State $\hat{\bm Y}_t^{\text{trend}} \gets \bm X_t^{\text{trend}} \cdot \bm W_t$
    \State $\hat{\bm Y}_t \gets \hat{\bm Y}_t^{\text{seasonal}} + \hat{\bm Y}_t^{\text{trend}}$
\Else
    \State $\hat{\bm Y}_t \gets \bm Z^{(d)} \cdot \bm W_o$
\EndIf
\State \Return $\hat{\bm Y}_t$
\end{algorithmic}
\end{algorithm}


\section{Toward Koopman-Augmented Transformer Architectures} \label{koopman}
The \textit{Koopman operator} provides a powerful, operator-theoretic framework to analyze nonlinear dynamical systems through the lens of linear dynamics in infinite-dimensional spaces. This operator acts on the space of observable functions, evolving them forward in time according to the system dynamics rather than directly propagating the states themselves \cite{Lusch2018,Yeung2019,NIPS2017_3a835d32}. This remarkable property enables the use of linear analysis techniques, even when the underlying system is nonlinear and potentially chaotic \cite{khosravi2023representer}. Recent developments have revitalized interest in Koopman theory, particularly through data-driven methods such as Dynamic Mode Decomposition (DMD), which offer practical means to approximate the operator from measurements and facilitate modal analysis, prediction, and control in complex systems~\cite{Drgona2022,Skomski2021,NEURIPS2021_c9dd73f5}.

While Koopman-based methods have traditionally been applied to deterministic systems governed by ordinary or partial differential equations—such as fluid dynamics, robotics, and chaotic systems—they are increasingly being explored for data-driven forecasting. Prior works such as \cite{lusch2018deep}, \cite{takeishi2017learning} demonstrated how deep networks can be used to approximate Koopman-invariant subspaces, enabling stable rollout predictions over time. However, most of these studies focus on dynamical reconstruction or control, rather than general-purpose time series forecasting.

Our approach bridges this gap by combining Koopman operator learning with Transformer-based sequence models, such as \patchtst, \autoformer, and \informer. We use a Transformer encoder to extract temporally-aware latent representations from input time series patches, and then apply a constrained Koopman operator in the latent space to evolve the system forward. The operator is spectrally regularized to ensure stability, and the decoder projects the evolved latent states back into the original observable space. While classical forecasting methods treat time series as stochastic sequences with limited structural assumptions, our hybrid framework explicitly enforces a structured latent dynamic, making it well-suited for forecasting systems that arise from physical or biological processes. To the best of our knowledge, this work is among the first to integrate Koopman theory with Transformer-based backbones for robust, interpretable multi-step forecasting.


\subsection{\dkoopformer~Building Blocks} \label{sec:building}

\dkoopformer~is a modular neural forecasting architecture that integrates the expressive power of Transformer encoders with the stability and interpretability of Koopman operator theory. Below, we outline the model components and their mathematical underpinnings.

\paragraph{Latent Representation via Transformer Encoder}
Given an observation $\bm x_t \in \mathbb{R}^{d}$ at time $t$, where $d$ denotes the number of observed variables, the encoder maps it into a latent representation:
\begin{equation}
    \bm z_t = \mathcal{E}_\theta(\bm x_t), \qquad \bm z_t \in \mathbb{R}^{d_{\mathrm{latent}}},
    \label{eq:latent}
\end{equation}
where $\mathcal{E}$ denotes encoder and $\theta$ are its associated parameters. The attention-based structure captures both short- and long-range dependencies in the input sequence without assuming periodic structure~\cite{vaswani2017attention}. In theory, the latent space dimension used by Koopman can differ from the embedding space of the Transformer (i.e. $d_{\mathrm{latent}}\ne d_{\mathrm{model}}$).


\paragraph{Koopman Latent Propagation}

In place of complex nonlinear transitions, \dkoopformer~models the temporal
evolution in latent space with a linear Koopman map:
\begin{equation}
    \bm z_{t+1} = \bm K\,\bm z_{t},
    \label{eq:K_evolve}
\end{equation}
where the operator is factorised as an orthogonal–diagonal–orthogonal (ODO)
product
\begin{align}
    \bm K &= \bm U\,\mathrm{diag}(\bm\Sigma)\,\bm V^{\!\top},\\
    \bm U^{\!\top}\bm U&=\bm I,\;
    \bm V^{\!\top}\bm V=\bm I,\\
    \Sigma_i&=\sigma(\Sigma_i^{\text{raw}})\,\rho_{\max},\;
    \rho_{\max}<1.
    \label{eq:odo}
\end{align}
Because every singular value is clipped below $\rho_{\max}$ we obtain the
operator-norm bound
\begin{equation}\label{bounded_eignvalue}
\lVert\bm K\rVert_2 \;=\; \max_i \Sigma_i \;\le\; \rho_{\max} \;<\; 1,
\end{equation}
which already guarantees contractive latent dynamics and bounded gradients
\footnote{If a strictly normal operator is required, one may simply tie the factors
by enforcing $\bm U=\bm V$, but this is not necessary for stability.}.

\paragraph{Decoder and End-to-End Forecasting}

The decoded prediction is generated through a linear map:
\begin{equation}
    \hat{\bm x}_{t+1} = \mathcal{D}_\phi(\bm z_{t+1}) := \bm W \bm z_{t+1}, \quad \bm W \in \mathbb{R}^{d \times d_{\mathrm{latent}}},
    \label{eq:decoder}
\end{equation}
where $\phi$ are decoder parameters. This leads to the overall differentiable composition:
\begin{equation}
    \hat{\bm x}_{t+1} = \mathcal{D}_\phi\left( \bm K \, \mathcal{E}_\theta(\bm x_t) \right).
    \label{eq:end2end}
\end{equation}

\subsection{Stability Guarantees and Lyapunov Convergence}
\label{sec:stability}

The bounded spectral radius $\rho(\bm K) < 1$ implies asymptotic contraction of latent dynamics:
\begin{equation}
    \| \bm K^n \|_2 \le \rho_{\max}^n, \qquad \lim_{n \to \infty} \bm K^n = \bm 0.
    \label{eq:K_decay}
\end{equation}
To further ensure energy decay in transients, we introduce a Lyapunov regularization term:
\begin{equation}
    L_{\mathrm{Lyap}} = \lambda \cdot \mathrm{ReLU} \left( \|\bm z_{t+1}\|^2 - \|\bm z_t\|^2 \right), \qquad \lambda > 0.
    \label{eq:lyap}
\end{equation}

While $\rho(\bm K)<1$ already guarantees asymptotic stability,
minimising $L_{\mathrm{Lyap}}$ expedites convergence during training by
dampening any residual transient growth~\cite{rodriguez2022lyanet,cui2021lyapunov,zheng2022lyapunov}.



\begin{corollary}[Global exponential stability]\label{cor:global_stab}
Under the dynamics \(\mathbf z_{t+1}=\bm K\mathbf z_t\) and the condition of \eqref{bounded_eignvalue},
\[
\|\bm z_{t+h}\|_2\;\le\;\rho_{\max}^{\,h}\|\bm z_t\|_2,
\qquad\text{so}\quad
\lim_{t\to\infty}\|\bm z_t\|_2=0.
\]
\end{corollary}

\begin{proposition}[Per-step contractivity]\label{prop:lyap}
If, in addition, the Lyapunov penalty
\(L_{\mathrm{Lyap}} = \lambda\,\mathrm{ReLU}\!\bigl(\|\mathbf z_{t+1}\|^2-\|\mathbf z_t\|^2\bigr)\)
is minimised to zero during training, every realised trajectory satisfies
\(
\|\bm z_{t+1}\|\le\|\bm z_t\|,
\)
providing a stronger sample-wise contraction.

\begin{proof}
Let $V(\bm z) = \|\bm z\|^2$ be the candidate Lyapunov function. Under the dynamics $\bm z_{t+1} = \bm K \bm z_t$, we compute the derivative:
\[
\Delta V := V(\bm z_{t+1}) - V(\bm z_t) = \bm z_t^\top (\bm K^\top \bm K - \bm I) \bm z_t.
\]
Since $\rho(\bm K) < 1$, we have $\bm K^\top \bm K - \bm I \prec 0$, which ensures $\Delta V < 0$ for all $\bm z_t \neq 0$. Thus, $V(\bm z_t)$ is strictly decreasing.

If $L_{\mathrm{Lyap}} = 0$, then $\Delta V \le 0$, and so the sequence $\{ V(\bm z_t) \}_{t=0}^\infty$ is monotone non-increasing and bounded below by zero. Hence, it converges:
\[
\lim_{t \to \infty} V(\bm z_t) = c \ge 0.
\]

Using the fact that $\|\bm z_{t+h}\| \le \|\bm K^h\|_2 \cdot \|\bm z_t\| \le \rho_{\max}^h \cdot \|\bm z_t\|$ for some $\rho_{\max} < 1$, we obtain:
\[
\lim_{h \to \infty} \|\bm z_{t+h}\| = 0,
\]
and therefore $V(\bm z_t) \to 0$. Thus, the origin is globally asymptotically stable.
\end{proof}

\end{proposition}

\vspace{0.5em}
\begin{corollary}[Robustness to Latent Perturbations]
Let $\bm z_t$ and $\tilde{\bm z}_t$ denote latent trajectories under clean and perturbed inputs, respectively. Assume $\rho(\bm K) < 1$ and $L_{\mathrm{Lyap}} = 0$. Then the deviation in predicted outputs satisfies:
\[
\|\hat{\bm x}_{t+h} - \tilde{\bm x}_{t+h}\| \le \|\bm W\|_2 \cdot \rho_{\max}^h \cdot \|\bm z_t - \tilde{\bm z}_t\|.
\]
Hence, prediction errors decay exponentially with forecast horizon $h$.
\end{corollary}

\begin{proof}
Let $\bm z_{t+h} = \bm K^h \bm z_t$ and $\tilde{\bm z}_{t+h} = \bm K^h \tilde{\bm z}_t$ be the propagated latents. Then the difference between predicted outputs is:
\[
\hat{\bm x}_{t+h} - \tilde{\bm x}_{t+h} = \bm W (\bm z_{t+h} - \tilde{\bm z}_{t+h}) = \bm W \bm K^h (\bm z_t - \tilde{\bm z}_t).
\]
Taking the 2-norm and using submultiplicativity:

\begin{multline}\label{eq:error_bound}
\|\hat{\bm x}_{t+h} - \tilde{\bm x}_{t+h}\| \le \|\bm W\|_2 \cdot \|\bm K^h\|_2 \cdot \|\bm z_t - \tilde{\bm z}_t\| \\ \le \|\bm W\|_2 \cdot \rho_{\max}^h \cdot \|\bm z_t - \tilde{\bm z}_t\|. 
\end{multline}
This shows that the model is exponentially stable under input noise or perturbations in the latent state.
\end{proof}


\begin{remark}[Spectral–Norm Clipping via an ODO Factorisation]
With the parameterisation  
\[
\bm K \;=\; \mathbf U\,\mathrm{diag}(\boldsymbol\Sigma)\,\mathbf V^{\!\top},\qquad
\Sigma_i = \sigma\!\bigl(\Sigma^{\mathrm{raw}}_i\bigr)\,\rho_{\max},\;0<\rho_{\max}<1,
\]
the induced operator norm satisfies  
\(
\|\bm K\|_{2}=\max_i\Sigma_i\le\rho_{\max}<1.
\)
Consequently every $h$-step map obeys the exponential bound  
\[
\|\bm K^{h}\|_{2}\le\rho_{\max}^{h},
\qquad
\Bigl\Vert\frac{\partial\mathbf z_{t+h}}{\partial\mathbf z_t}\Bigr\Vert_{2}
      = \|\bm K^{h}\|_{2}\le\rho_{\max}^{h},
\]
so the {\em forward dynamics are contractive} and {\em back-propagated gradients are uniformly bounded} across all rollout horizons.  Empirical evidence for the effectiveness of spectrum clipping and related singular-value control can be found in \cite{fan2024learning,guo2024spectrum,ebrahimpour2024spectrum}.
\end{remark}

\begin{remark}[Per-step contractivity enforced by $L_{\mathrm{Lyap}}$]
When the Lyapunov penalty is driven to zero during training, every {\em realised}
latent trajectory satisfies the one–step energy inequality
\[
\|\bm z_{t+1}\| \;\le\; \|\bm z_t\|\qquad\text{for all training samples.}
\]
Hence the sequence $\bigl\{\|\bm z_{t+h}\|\bigr\}_{h\ge 0}$ is {\em monotonically
non-increasing}—a stronger guarantee than the
spectral-radius bound $\|\bm K^h\|_2\le\rho_{\max}^h$, which only constrains the
{\em worst-case} direction.  In practice this per-step contractivity prevents
early-stage gradient explosions and shortens the transient tail, yielding faster
and more stable training \cite{cui2021lyapunov, zinage2023neural, dawson2023safe, dai2021lyapunov}. 
\end{remark}


\subsection{Empirical Implications and Gradient Stability}

The theoretical guarantees in the previous theorem translate into several empirical advantages during training and deployment of \dkoopformer:

\paragraph{Gradient Stability via Spectral Boundedness}
The Koopman operator $\bm K$ is parameterized via orthogonal–diagonal–orthogonal (ODO) factorization:
\[
\bm K = \bm U \cdot \mathrm{diag}(\bm \Sigma) \cdot \bm V^\top,
\]
where $\bm U$, $\bm V$ are orthogonal matrices, and $\bm \Sigma_i = \sigma(\Sigma_i^{\mathrm{raw}}) \cdot \rho_{\max}$ with $\rho_{\max} < 1$. This design ensures:
\begin{itemize}
    \item $\|\bm K\|_2 = \max_i \Sigma_i \le \rho_{\max}<1$,
    
    \item The sigmoid clip gives $|\partial\sigma/\partial\Sigma_i^{\mathrm{raw}}| \le 1/4$. Hence every gradient \(\bigl\lVert \partial\hat{\bm x}_{t+1}/\partial\Sigma_i^{\mathrm{raw}}\bigr\rVert
\le \lVert\bm W\rVert_2\,\rho_{\max}/4\) is uniformly bounded.
\end{itemize}
These properties control spectral radius during training and prevent gradient explosion.

\paragraph{Stable Backpropagation through Koopman Layers}
Since $\bm K$ is differentiable with respect to $\{\bm U, \bm V, \bm \Sigma^{\mathrm{raw}}\}$ and is spectrally normalized, the gradient of the end-to-end forecast
\[
\hat{\bm x}_{t+1} = \bm W \cdot \bm K \cdot \bm z_t
\]
with respect to any component of $\bm K$ is uniformly bounded:
\[
\left\| \frac{\partial \hat{\bm x}_{t+1}}{\partial \Sigma_i^{\mathrm{raw}}} \right\| \le \|\bm W\|_2 \cdot \|\bm U\|_2 \cdot \|\bm V\|_2 \cdot \rho_{\max} \cdot \frac{1}{4},
\]
where the derivative of the sigmoid is upper bounded by $1/4$. Hence, $\nabla_{\bm \Sigma^{\mathrm{raw}}} L$ is numerically well-behaved.

\paragraph{No Exploding Activations in Latent Dynamics}
Thanks to the Lyapunov regularization $L_{\mathrm{Lyap}}$, the model actively penalizes transient amplification. This results in:
\[
\max_t \|\bm z_t\|^2 \le \|\bm z_0\|^2,
\]
which bounds the norm of intermediate representations even under long forecast horizons or recursive unrolling.

\paragraph{Robust Generalization with Long Horizons}
The exponential error decay property,
\[
\|\hat{\bm x}_{t+h} - \tilde{\bm x}_{t+h}\| \le \|\bm W\|_2 \cdot \rho_{\max}^h \cdot \|\bm z_t - \tilde{\bm z}_t\|,
\]
prevents cumulative error explosion common in autoregressive models. Empirically, this enhances:
\begin{itemize}
    \item Forecast quality on long-horizon benchmarks,
    \item Robustness to sensor noise and distribution shift,
    \item Stability under recursive forecasting or closed-loop rollout.
\end{itemize}

\paragraph{Numerical Conditioning via Orthogonality}

By re-projecting $\bm U$ and $\bm V$ onto the Stiefel manifold---that is, the space of matrices with orthonormal columns---using Householder QR, we ensure that both $\bm U^\top \bm U = \bm I$ and $\bm V^\top \bm V = \bm I$ hold exactly during training, maintaining numerical stability and avoiding drift in the eigenvector bases. Therefore, we guarantee:

\begin{align*}
\mathrm{cond}(\bm U) &= \mathrm{cond}(\bm V) = 1, \quad
\text{and, provided } \Sigma_i>0,\;\\
\kappa(\bm K)&=\frac{\max_i \Sigma_i}{\min_i \Sigma_i}
\le \rho_{\max}\big/\min_i \Sigma_i,
\end{align*}
where, $\operatorname{cond}(\mathbf U=\kappa_2(\mathbf V):=\|\mathbf U\|_2\,\|\mathbf U^{-1}\|_2=\sigma_{\max}(\mathbf U)/\sigma_{\min}(\mathbf U)$ denotes the spectral-norm (2-norm) condition number, i.e.\ the factor by which the worst-case relative error in a vector can be amplified when it is transformed by $\mathbf U$. This prevents numerical drift in eigenvector bases, which is critical when training models with multiple Koopman propagation steps.

\paragraph{Comparison with Prior Work} \label{sec:compare}

\vspace{0.5em}
\noindent Table~\ref{tab:matrix} summarizes key innovations of \dkoopformer~compared to related Koopman-inspired forecasting models.

\begin{table*}
    \centering
    \small
    \begin{tabular}{@{}lcccc@{}}
        \toprule
        \textbf{Property} &
        \textbf{\dkoopformer} &
        DeepKoopman \cite{lusch2018deep} &
        Koopman Neural Forcaster (KNF) \cite{wang2022koopman} &
        Koopformer \cite{wang2023koopformer} \\
        \midrule
        Strict $\rho(\bm K) < 1$        & \checkmark & -- & -- & -- \\
        Lyapunov dissipation            & \checkmark & -- & -- & -- \\
        Normal operator                 & \checkmark & -- & -- & -- \\
        Encoder plug-and-play           & \checkmark & MLP & RNN & Custom Trf \\
        Provable error bound            & Eq.~\eqref{eq:error_bound} & -- & -- & -- \\
        Orthogonal factors              & QR (exact) & -- & -- & Cayley
 \\
        \bottomrule
    \end{tabular}
    \caption{Comparison of \dkoopformer~ with related Koopman-based forecasting models. \emph{Cayley} refers to the parameterization of orthogonal factors via the Cayley transform $\bm Q = (\bm I - \frac{1}{2} \bm A)^{-1} (\bm I + \frac{1}{2} \bm A)$ with a skew-symmetric matrix $\bm A$, yielding approximate (but not exact) orthogonality~\cite{wang2023koopformer}. \emph{Custom Trf} indicates that Koopformer employs a bespoke Transformer encoder architecture with Koopman-inspired modifications, in contrast to the plug-and-play encoder design of \dkoopformer.}
    \label{tab:matrix}
\end{table*}



\begin{algorithm}
\caption{\dkoopformer~for Multivariate Time Series Forecasting}
\label{alg:koopformer}
\textbf{Input:} Multivariate time series $\{\bm x_t\}_{t=1}^T$, context length $P$, forecast horizon $H$\\
\textbf{Output:} Predicted sequence $\hat{\bm Y}_t = [\hat{\bm x}_{t+P}, \dots, \hat{\bm x}_{t+P+H-1}]$

\begin{algorithmic}[1]
\State \textbf{Data Preparation:}
    \begin{itemize}
        \item For each valid $t$, construct:
            \begin{align*}
                \bm X_t &= [\bm x_t, \dots, \bm x_{t+P-1}] \in \mathbb{R}^{P \times d}, \\
                \bm Y_t &= [\bm x_{t+P}, \dots, \bm x_{t+P+H-1}] \in \mathbb{R}^{H \times d}.
            \end{align*}
    \end{itemize}

\State \textbf{Latent Encoding:}
    \begin{itemize}
        \item Compute latent vector: $\bm z_t = \mathcal{E}_\theta(\bm X_t)$, where $\mathcal{E}_\theta$ is a Transformer-based encoder with positional encoding.
    \end{itemize}

\State \textbf{Koopman Operator Evolution:}
    \begin{itemize}
        \item Propagate the latent state:
            \[
            \bm z_{t+1} = \bm K \bm z_t.
            \]
        \item $\bm K$ is parameterized as:
            \[
            \bm K = \bm U\,\mathrm{diag}(\bm \Sigma)\,\bm V^\top.
            \]
            where $\bm U, \bm V$ are orthogonal matrices, $\bm \Sigma = \sigmoid(\bm \Sigma^{\mathrm{raw}})\,\rho_{\max}$, and $\rho_{\max} < 1$.
    \end{itemize}

\State \textbf{Output Decoding:}
    \begin{itemize}
        \item Forecast next $H$ steps by recursively propagating and decoding:
            \begin{align*}
                \text{for } h &= 1 \text{ to } H:\\ \quad
                &\bm z_{t+h} = \bm K \bm z_{t+h-1}, \\
                &\hat{\bm x}_{t+P+h-1} = \mathcal{D}_\phi(\bm z_{t+h}).
            \end{align*}
        \item Collect $\hat{\bm Y}_t = [\hat{\bm x}_{t+P}, \dots, \hat{\bm x}_{t+P+H-1}]$
    \end{itemize}

\State \textbf{Loss Function:}
    \begin{align*}
        \mathcal{L} = \|\hat{\bm Y}_t - \bm Y_t\|^2 + \lambda \cdot \mathrm{ReLU}\left(\|\bm z_{t+1}\|^2 - \|\bm z_t\|^2\right).
    \end{align*}

\State \textbf{Optimization:} Train parameters $(\theta, \phi, \bm K)$ end-to-end with Adam.

\State \textbf{Inference:} For long-term prediction, unroll Koopman propagation recursively:
    \[
    \bm z_{t+h} = \bm K^h \bm z_t, \quad \hat{\bm x}_{t+P+h-1} = \mathcal{D}_\phi(\bm z_{t+h}).
    \]
\end{algorithmic}
\end{algorithm}


\section{Numerical Simulations}\label{sec:simulations}

To assess the effectiveness of the proposed \dkoopformer~architecture across a diverse range of dynamical regimes, we conduct a comprehensive suite of numerical experiments. These experiments cover both synthetic and real-world time series forecasting tasks, designed to systematically evaluate the accuracy, stability, and generalization capabilities of Koopman-enhanced Transformer variants. We begin with controlled studies on noisy nonlinear dynamical systems, including the Van der Pol and Lorenz systems, where the ground truth dynamics are well understood. We then extend our evaluation to high-dimensional real-world datasets—including CMIP6-based climate projections, ERA5 reanalysis data, a cryptocurrency financial time series, and electricity generation records—to benchmark the models on realistic forecasting challenges involving multivariate, domain-specific, and spatially distributed time series.

Across all experiments, we compare \dkoopformer~variants (\patchtst, \autoformer, and \informer) against a conventional LSTM baseline under unified training conditions. Our evaluation spans multiple forecasting horizons, latent dimensions, and patch configurations, providing a detailed performance analysis under varying model capacities and temporal resolutions. To reduce parameter complexity, we set $d_{\mathrm{latent}} = d_{\mathrm{model}}$ throughout our experiments. However, the two quantities remain conceptually distinct: $d_{\mathrm{model}}$ governs Transformer representation capacity, while $d_{\mathrm{latent}}$ defines the dimensionality of the Koopman-invariant subspace.

\subsection{Hardware Setup}
All deep learning model training were performed on a high-performance computing node equipped with an \texttt{NVIDIA A100} \texttt{GPU} featuring $80$~GB of high-bandwidth Video RAM (\texttt{VRAM}), optimized for large-scale machine learning workloads. The system was configured with $256$~\texttt{GB} of main memory accessible per \texttt{CPU} core, ensuring efficient handling of high-dimensional data and large batch sizes. Training tasks were orchestrated using the Simple \texttt{Linux} Utility for Resource Management (\texttt{SLURM}), which allocated compute resources with job-specific constraints to ensure optimal utilization of the \texttt{A100}'s memory architecture. This hardware-software configuration significantly accelerated the training process and maintained computational stability, which was critical for learning long-range temporal dependencies in climate time series data.

\paragraph{Code availability statement}
The \dkoopformer~framework---including source code, datasets, and representative figures---is accessible at \texttt{Github} \footnote{\url{https://github.com/Ali-Forootani/deepkoopformer}} and \texttt{Zenodo} \footnote{\url{https://doi.org/10.5281/zenodo.15826887, https://doi.org/10.5281/zenodo.15828000}}. The entire implementation is written in \texttt{Python} and builds upon standard scientific computing libraries, including \texttt{PyTorch}, \texttt{NumPy}, and \texttt{SciPy}.


\subsection{Application to Noisy Dynamical Systems}

An important application area for Koopman-enhanced Transformer architectures is the forecasting of noisy nonlinear dynamical systems. As demonstrated in this study with synthetic signals, real-world processes often exhibit complex temporal dynamics combined with stochastic perturbations.

\paragraph{Van der Pol System}
The Van der Pol system is a nonlinear second-order dynamical system that exhibits self-sustained oscillations and is widely used as a benchmark in nonlinear dynamics and control. It is defined by two coupled ordinary differential equations describing the evolution of the state vector $\mathbf{x}(t) = [x_1(t), x_2(t)]^\top$, given by:
\begin{align}
\frac{dx_1}{dt} &= x_2, \nonumber \\
\frac{dx_2}{dt} &= \mu (1 - x_1^2) x_2 - x_1, \nonumber
\end{align}
where $\mu > 0$ governs the nonlinearity and strength of damping. For small $\mu$, the system exhibits near-harmonic behavior, while for larger $\mu$, it produces relaxation oscillations with alternating fast and slow dynamics. In our simulations, we use $\mu = 1.0$, which leads to a nonlinear but stable oscillatory regime.

To simulate more realistic behavior under measurement uncertainty or environmental perturbations, we inject additive Gaussian noise into each state variable at every time step. The system is discretized using the Euler method with a fixed step size $\Delta t = 0.01$ over a time horizon of $T = 20.0$ seconds. At each step, zero-mean Gaussian noise with standard deviation $\sigma_n = 0.02$ is independently added to both $x_1$ and $x_2$. The resulting trajectory reflects the nonlinear oscillatory dynamics of the system while capturing stochastic variability, making it a robust testbed for system identification and forecasting models.

In our \dkoopformer~experiments on the Van der Pol system, all backbone architectures are configured for a fair and meaningful comparison under equivalent settings. The input sequence was segmented into patches of length $p=16$, and the forecast horizon is set to $H=5$. For all the \dkoopformer~variants, the latent dimension is set to $d_{\text{latent}}=d_{\text{model}}=16$, with $2$ attention heads and $2$ encoder layers. The Koopman operator layer is strictly stabilized by dynamically constraining its eigenvalues below $\rho (\bm K)= 0.99$ during training, ensuring stability and dissipativity in the learned linear dynamics. Models are trained using the Adam optimizer with a learning rate of $0.001$ for $1000$ epochs.

\begin{figure}
    \centering
    \includegraphics[width=0.99\linewidth]{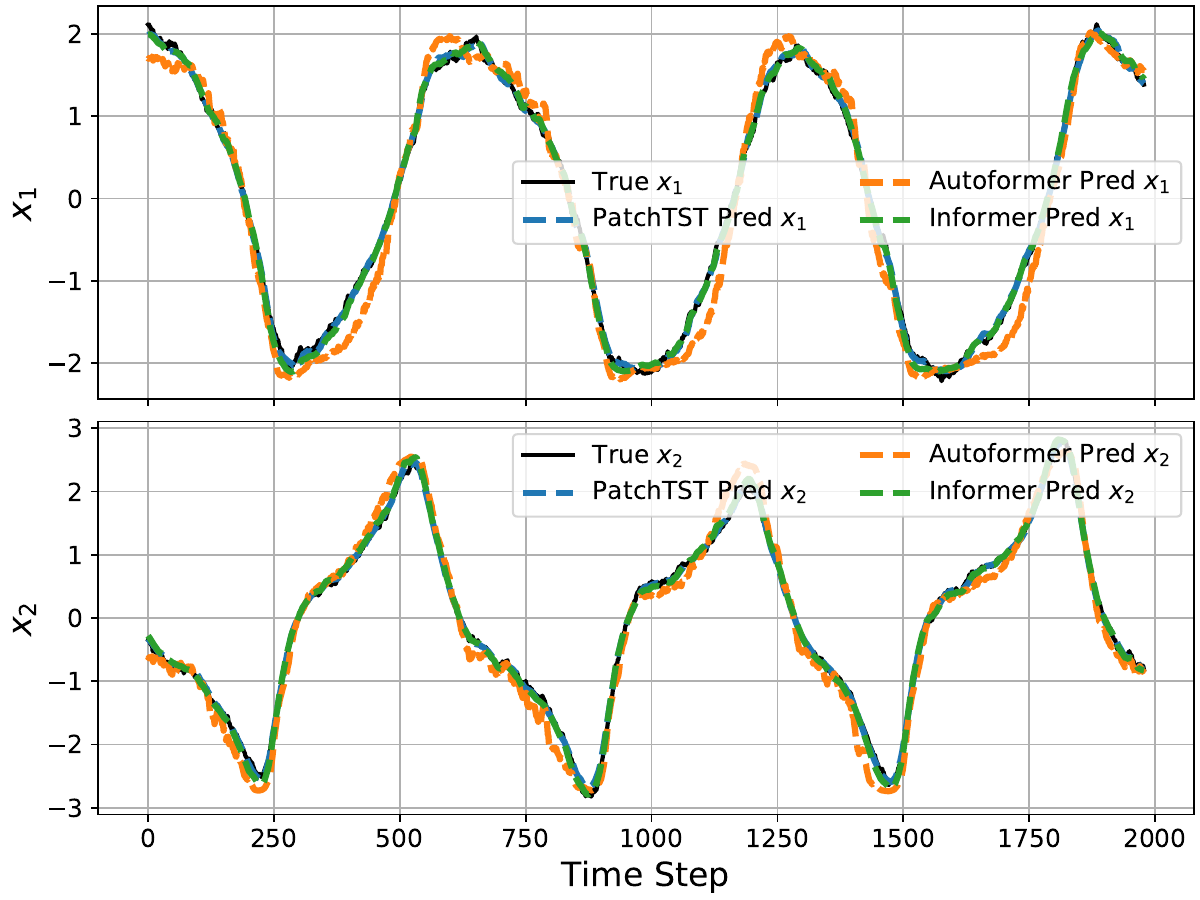}
    \caption{\dkoopformer~based Transformers variants comparison on the Van der Pol System.}
    \label{fig:van_der_pol}
\end{figure}

\begin{figure}
    \centering
    \includegraphics[width=0.75\linewidth]{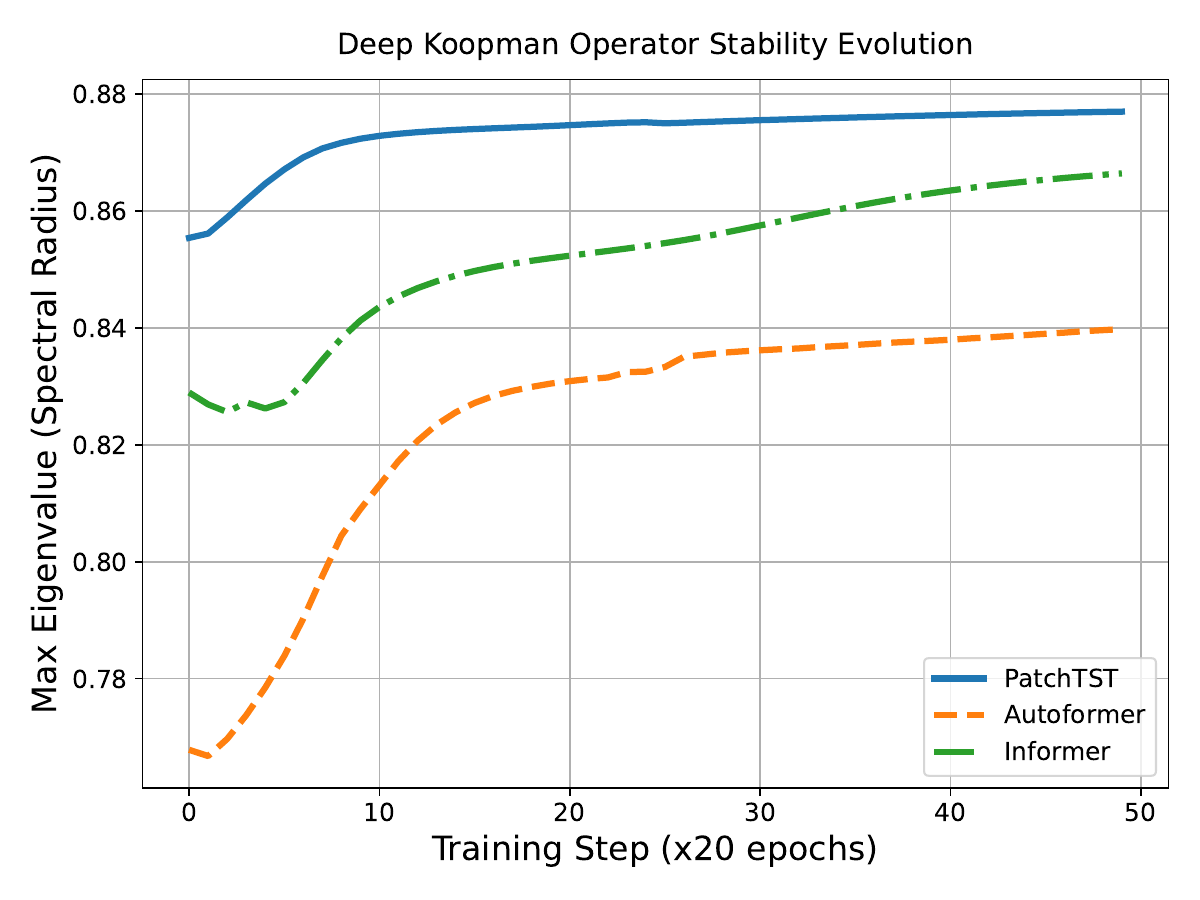}
    \caption{Comparison of \dkoopformer~operator stability evolution in variants Transformers on Van der Pol system during the training loop.}
    \label{fig:van_der_pol_eigenvalue}
\end{figure}

As shown in Fig.~\ref{fig:van_der_pol} and Fig.~\ref{fig:van_der_pol_eigenvalue}, the results from the training and evaluation indicate that all variants of the \dkoopformer~architecture perform similarly well in capturing the nonlinear oscillatory behavior of the Van der Pol system. They achieve accurate forecasting of both state variables, \(x_1\) and \(x_2\), while maintaining stable latent dynamics. This stability is further ensured by the Koopman operator, which constrains the evolution of the latent states, and is validated through the spectral radius of the operator during training (see Fig.~\ref{fig:van_der_pol_eigenvalue}). 


\paragraph{Lorenz System} The Lorenz system is a classical prototype of deterministic chaos, originally introduced to model atmospheric convection. It comprises three coupled nonlinear differential equations governing the evolution of the state vector \( \mathbf{x}(t) = [x_1(t), x_2(t), x_3(t)]^\top \), given by:
\begin{align}
\frac{dx_1}{dt} &= \sigma (x_2 - x_1), \nonumber \\
\frac{dx_2}{dt} &= x_1(\rho - x_3) - x_2, \nonumber \\
\frac{dx_3}{dt} &= x_1 x_2 - \beta x_3, \nonumber
\end{align}
where \( \sigma \), \( \rho \), and \( \beta \) are system parameters. In our experiments, we use the standard chaotic configuration: \( \sigma = 10.0 \), \( \rho = 28.0 \), and \( \beta = \tfrac{8}{3} \). To simulate a more realistic scenario with process uncertainty, we augment the system with additive Gaussian noise at each time step. The system is numerically integrated using the Euler method with a fixed time step \( \Delta t = 0.01 \) and total duration \( T = 20.0 \) seconds. Specifically, we add zero-mean Gaussian noise with standard deviation \( \sigma_n = 0.1 \) independently to each variable. The resulting trajectory forms a noisy, high-dimensional time series that captures both the chaotic geometry and the stochastic fluctuations inherent in real-world measurements. 

All \dkoopformer~variants are trained on Lorenz system data using a unified set of hyperparameters to ensure comparability. The patch length is set to $p= 150$, with a forecast horizon of $H= 5$ time steps. The backbone architectures (\patchtst, \autoformer, \informer) use a hidden (latent) dimension of $d_{\text{model}}=d_{\text{latent}}=16$ for all \dkoopformer~variants, reflecting their respective architectural capacities. Transformer encoder layers utilize $2$ attention heads and a feedforward dimension of $64$. Each backbone is coupled with a strictly stable Koopman operator layer, whose spectral radius is enforced to be less than $0.99$ via a sigmoid-based reparameterization. Training is conducted for $3000$ epochs using the Adam optimizer with a learning rate of $0.001$. The loss function consists of the mean squared error (MSE) between predicted and ground truth outputs, augmented by Lyapunov regularization term with $\lambda=0.1$ that penalizes the increment in the latent norm post-Koopman evolution. 

\begin{figure}
    \centering
    \includegraphics[width=0.99\linewidth]{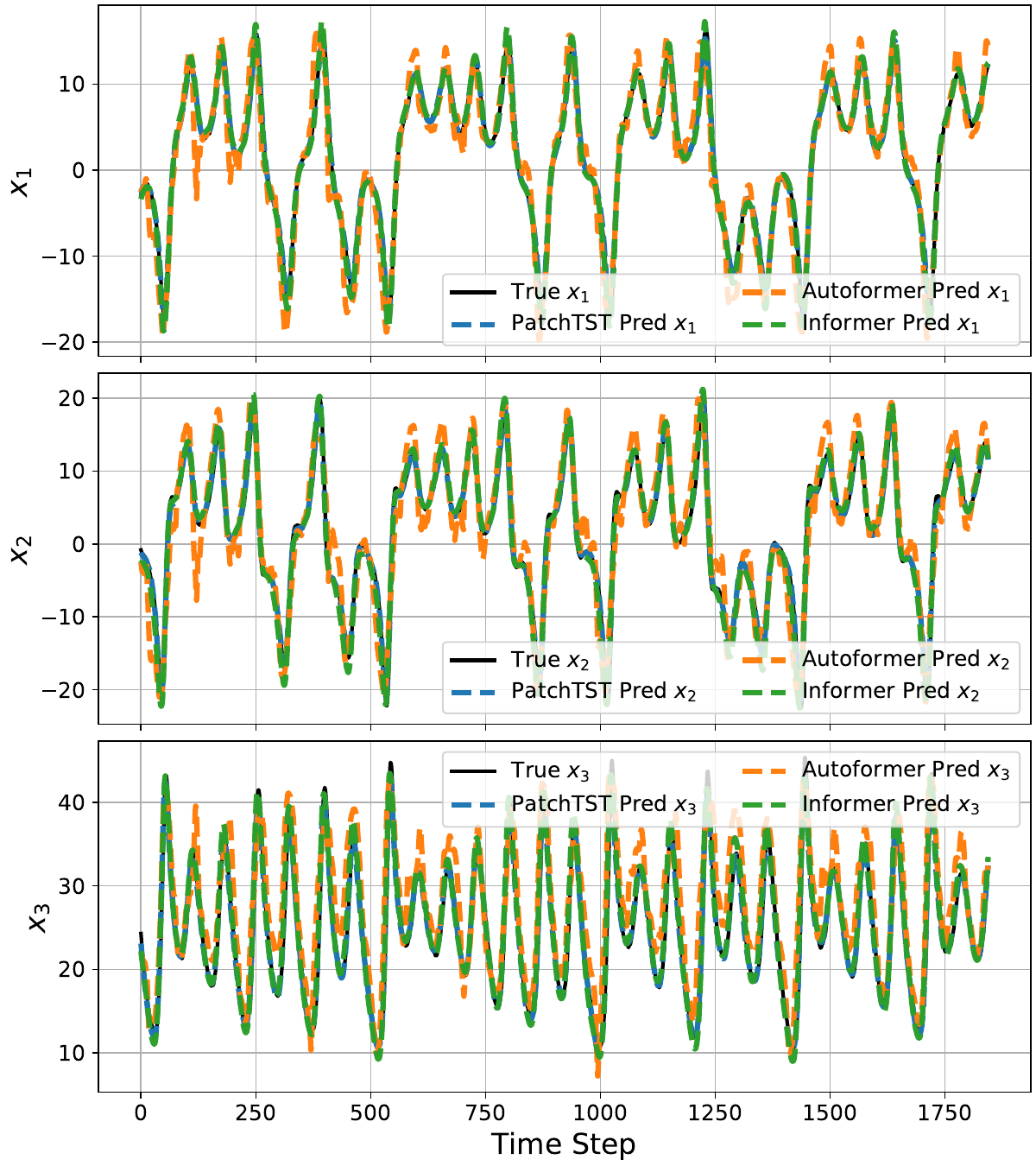}
    \caption{\dkoopformer~variant prediction on Lorenz dynamical system.}
    \label{fig:lorenz_system}
\end{figure}

\begin{figure}
    \centering
    \includegraphics[width=0.75\linewidth]{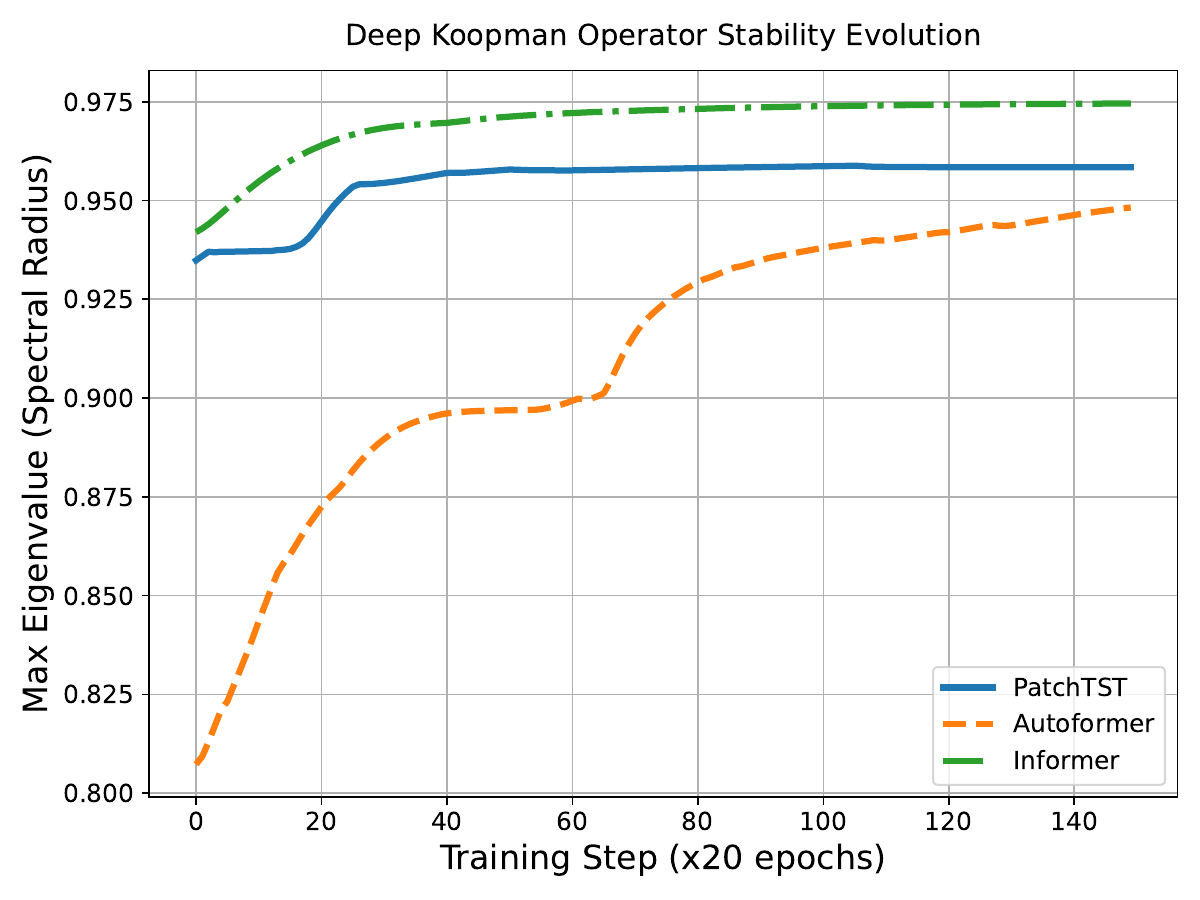}
    \caption{Comparison of \dkoopformer~operator stability evolution in variants Transformers on Lorenz system during the training loop.}
    \label{fig:lorenz_system_eigenvalue}
\end{figure}

Fig.~\ref{fig:lorenz_system} presents the multistep forecasting results of the \dkoopformer~variants (\patchtst, \autoformer, and \informer) on the noisy Lorenz system. The predicted trajectories for all three state variables ($x_1$, $x_2$, $x_3$) are shown against the ground truth. All models achieve a high degree of accuracy in reproducing the underlying chaotic dynamics, closely tracking the true trajectories over the entire test window. Notably, the \patchtst~and \informer~variants yield visually indistinguishable predictions from the ground truth, reflecting their strong temporal modeling capabilities. The \autoformer~model also demonstrates robust performance, although it exhibits slightly higher deviations during certain oscillatory regimes, particularly in the $x_2$ and $x_3$ states. Overall, these results confirm that Koopman-enhanced Transformer backbones are effective for noisy, nonlinear multivariate time series, with \patchtst~and \informer~providing marginally superior consistency in long-range prediction. Figure~\ref{fig:lorenz_system_eigenvalue} shows the evolution of the largest Koopman operator eigenvalue (spectral radius) during training for each \dkoopformer~variant, illustrating that all models enforce spectral stability while converging to slightly different asymptotic operator norms. Notably, the \informer~and \patchtst~variants stabilize near the prescribed upper bound, whereas the \autoformer~model converges more gradually over the course of training.

\subsection{Application of \dkoopformer~on climate dataset}

In this study, we employ climate data from the Coupled Model Intercomparison Project Phase~6 (CMIP6 \footnote{\url{https://cds.climate.copernicus.eu/datasets/projections-cordex-domains-single-levels?tab=overview}}) to analyze surface wind speed and atmospheric pressure over Germany~\cite{makula2022coupled}. CMIP6 provides standardized climate projections derived from multiple global climate models, offering a robust foundation for assessing regional meteorological patterns. In our previous study~\cite{forootani2024climate}, we adapted CMIP6 data for localized wind energy applications in Germany. This process involved spatial interpolation to transform the rotated pole grid into standard geographic coordinates, specifically targeting the locations of operational wind farms. Temporal resampling and feature scaling were subsequently applied to harmonize the data structure and enable effective use in deep learning models. The resulting dataset captures key atmospheric dynamics and supports high-resolution forecasting of wind power potential under evolving climate conditions~\cite{forootani2024climate}. Complementing this, we leverage the ERA5 \footnote{\url{https://cds.climate.copernicus.eu/datasets}} dataset~\cite{hersbach2020era5}, developed by the European Center for Medium-Range Weather Forecasts (ECMWF), which provides hourly estimates of numerous climate variables through global atmospheric reanalysis. ERA5 combines model outputs with observations via advanced data assimilation techniques, offering high spatial resolution (approximately 31~km) and extensive temporal coverage (1950 to present). Its variables---including wind speed, surface pressure, temperature, and radiation fluxes---serve as a reliable reference for validating CMIP6-based projections and for benchmarking the forecasting performance of our \dkoopformer~models. In this section, we evaluate \dkoopformer~forecasting capabilities on wind speed and surface pressure data sourced from both CMIP6 and ERA5.

We systematically benchmark \dkoopformer~for multi-step wind speed forecasting, including the Koopman-augmented \patchtst, \autoformer, and \informer~variants, alongside a standard LSTM baseline. As explained in the \dkoopformer~models, each network is structured into three principal modules: (i) a deep encoder backbone (\patchtst, \autoformer, or \informer), which processes the input sequence to extract latent representations; (ii) a linear Koopman operator, which advances the latent state in time, imposing stable, linear evolution dynamics in the learned feature space; and (iii) a linear decoder, which reconstructs the forecasted outputs from the propagated latent state. Specifically, the \patchtst~backbone employs temporal patch embedding and multi-head self-attention, the \autoformer~backbone integrates time-series decomposition for explicit trend and seasonal modeling, and the \informer~backbone uses efficient sparse self-attention mechanisms. The Koopman operator itself is parameterized via an orthogonal-diagonal-orthogonal (ODO) factorization with spectral constraints to ensure stability.

Key hyperparameters include the latent (hidden) dimension ($d_{\mathrm{model}}$), which controls the width of the latent feature space; the context window length ($P$), representing the number of historical steps provided to the model; the forecast horizon ($H$), which sets the number of steps predicted into the future; and the patch length ($p$), which defines the size of non-overlapping temporal patches (for \patchtst, \informer, and \autoformer).

Data is standardized using \text{MinMax} scaling. All models are trained and evaluated under the same random seed and data splits to ensure strict reproducibility and fair comparison. 
This configuration allows systematic comparison of network capacity, stability, and predictive accuracy across Koopman-augmented (\dkoopformer) and conventional neural (LSTM) forecasting architectures.

Model parameters are optimized using the Adam optimizer with a learning rate of $3 \times 10^{-4}$, and trained for up to $4000$ epochs per configuration.

In all Koopformer-based models, the Koopman operator was constrained to have a spectral radius less than $\rho(\bm K) = 0.99$ and included Lyapunov stability regularization ($\lambda=0.1$) to promote stable latent dynamics. The same spatial indices were used for each trial, enabling direct per-channel and per-configuration evaluation.

\subsubsection{Scenario 1}\label{scenario_1}
In the first scenario for all transformer-based models, the encoder comprises $3$ layers and $4$ attention heads, with a hidden dimension of $d_{\text{model}}=d_{\text{latent}}= 48$ and a feedforward width of $96$.

The LSTM baseline consists of two layers with a hidden size of $64$. Model parameters are optimized using the Adam optimizer with a learning rate of $3\times 10^{-4}$, and training is performed for up to $4000$ epochs. Grid searches are conducted across a range of patch lengths ($p \in \{80, 90, 100, 110, 120, 130\}$) and forecast horizons ($H \in \{10, 15, 20, 25, 30\}$), with $5$ representative input features ($5$ channel) per sample. 

\begin{figure*}
    \centering
    \includegraphics[width=0.70\linewidth]{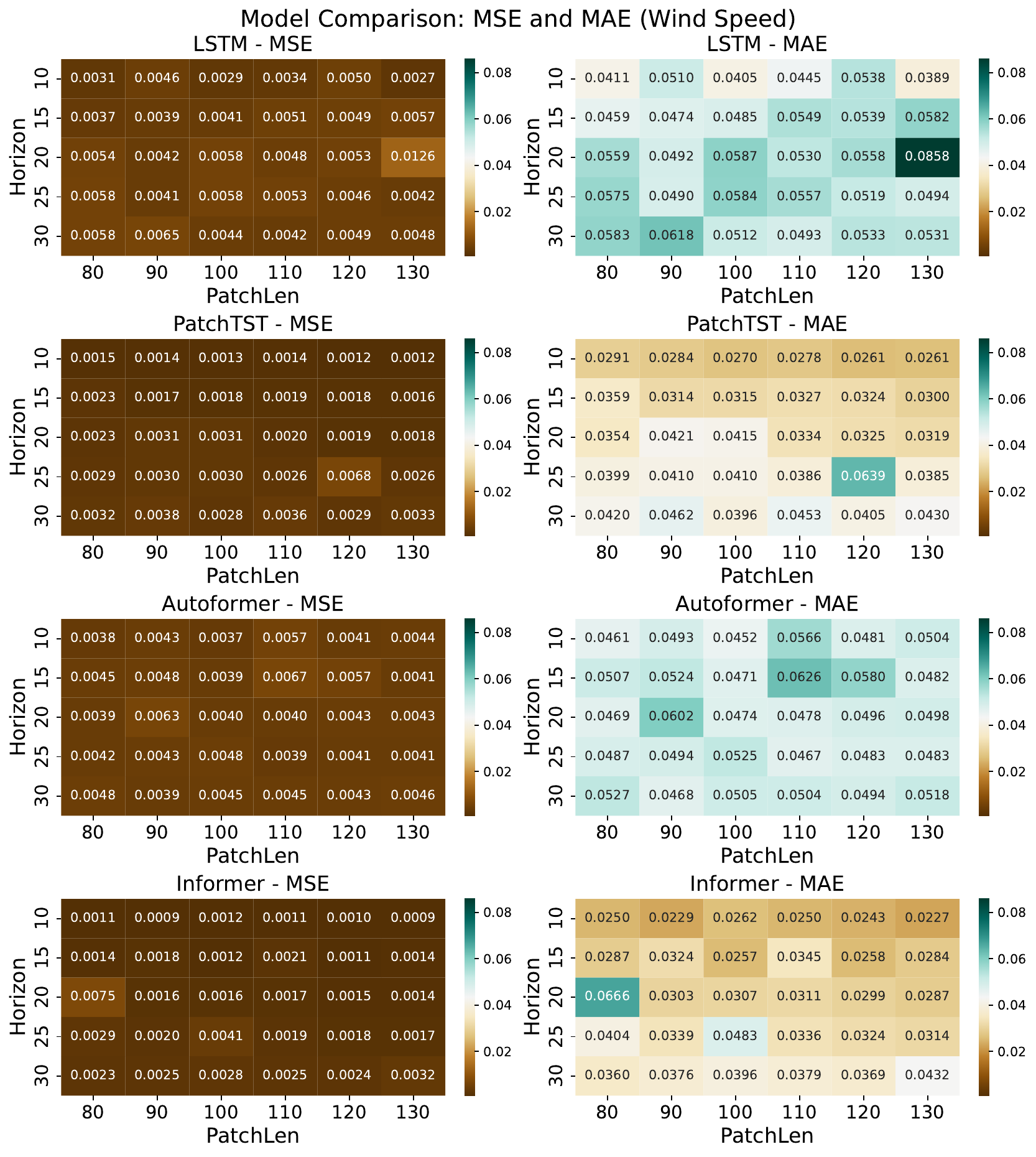}
    \caption{\dkoopformer versus LSTM baseline grid search for in CMIP6 dataset wind speed forecasting in scenario 1.}
    \label{fig:wind_speed_fixed_dmodel}
\end{figure*}

The benchmarking results, summarized in Fig. \ref{fig:wind_speed_fixed_dmodel}, reveal a consistent performance advantage of the Koopman-augmented \patchtst~and \informer~models over the standard LSTM baseline for wind speed forecasting. Across all combinations of patch length and forecast horizon, both \patchtst~and \informer~achieve markedly lower Mean Squared Error (MSE) and Mean Absolute Error (MAE) values compared to LSTM. This performance gap is particularly pronounced at longer forecast horizons, where the LSTM's errors increase significantly, while the Koopman-augmented transformer models maintain robust predictive accuracy. Notably, the \patchtst~and \informer~variants demonstrate remarkable stability with respect to the patch length hyperparameter, showing only minor variations in error across the tested range. In contrast, the LSTM—lacking explicit patching or attention mechanisms—exhibits higher sensitivity to forecast horizon and overall higher error rates. The \autoformer~model also outperforms the LSTM, but generally lags slightly behind \patchtst~and \informer~in this task. These findings strongly support the use of transformer-based, Koopman-augmented architectures for multi-step time series forecasting, as they combine powerful temporal abstraction with stable and generalizable latent dynamics, resulting in superior accuracy and robustness compared to traditional recurrent models.

\begin{figure}
    \centering
    \includegraphics[width=0.99\linewidth]{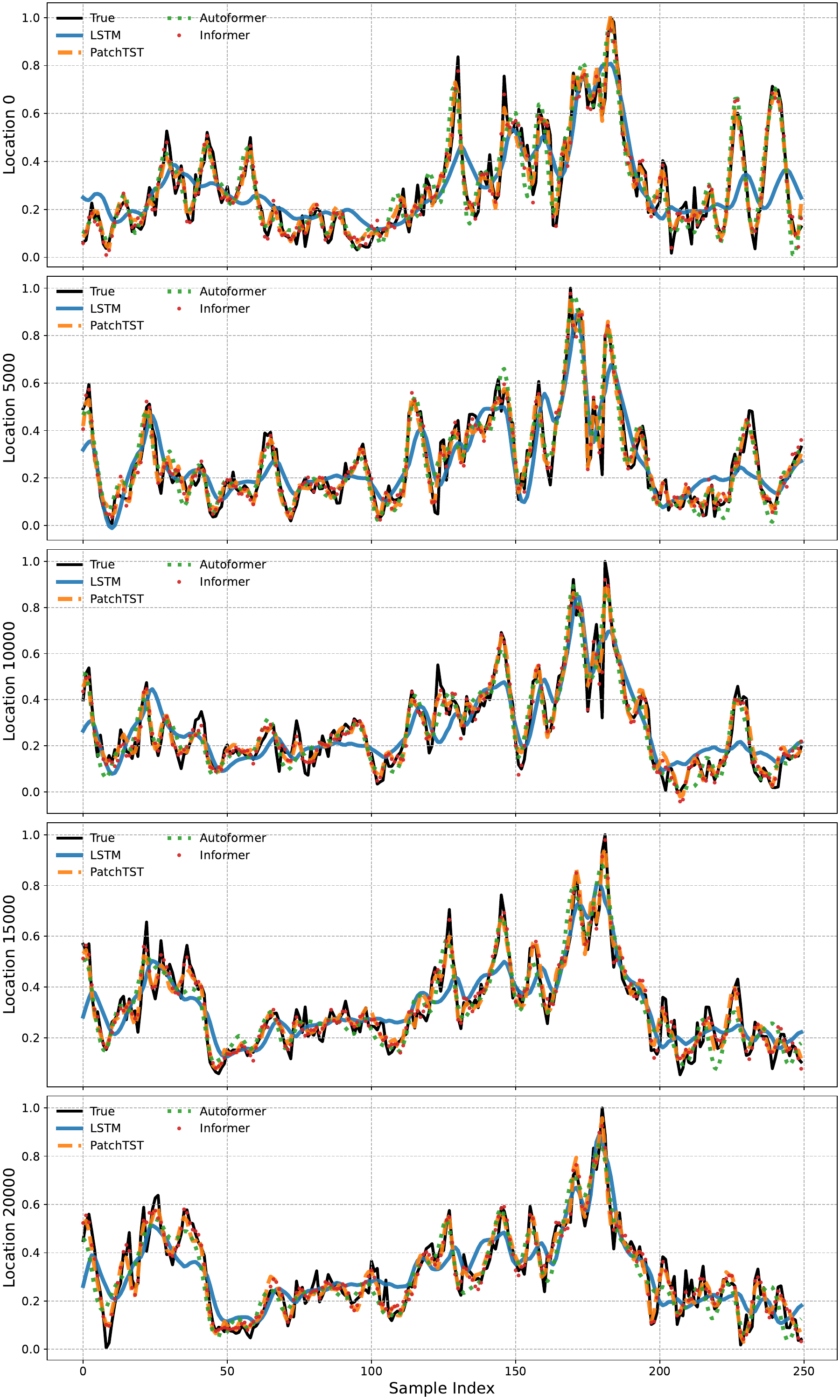}
    \caption{Comparison of \dkoopformer~with LSTM approach in wind speed forecasting with patch length $p=130$, horizon $H=20$, and $d_{\text{model}}=48$.}
    \label{fig:wind_speed_fixed_dmodel_1setting}
\end{figure}

The qualitative forecast trajectories in Fig. \ref{fig:wind_speed_fixed_dmodel_1setting} further support the quantitative error results: \dkoopformer~models closely reproduce the amplitude, phase, and fine-scale dynamics of the true wind speed across all $5$ spatial locations, regardless of underlying signal complexity. In contrast, the LSTM baseline, while able to capture the gross trend, systematically underfits sharp transitions and exhibits smoothed, lagging forecasts. These findings reinforce the superiority of attention-based, \dkoopformer~architectures for high-fidelity, multi-channel wind speed forecasting in spatiotemporal settings.

\subsubsection{Scenario 2}
In the second scenario, we conduct grid sweeps over $d_{\mathrm{model}} \in \{8, 16, 24, 32, 40, 48\}$ and forecast horizons $H \in \{10, 15, 20, 25, 30\}$. All transformer-based backbones use $3$ encoder layers, $4$ attention heads, sinusoidal positional encodings, and a feedforward network width of $96$. The LSTM baseline consists of $2$ stacked layers with a hidden size of $d_{\text{model}}=\{8, 16, 24, 32, 40, 48\}$.

\begin{figure*}
    \centering
    \includegraphics[width=0.70\linewidth]{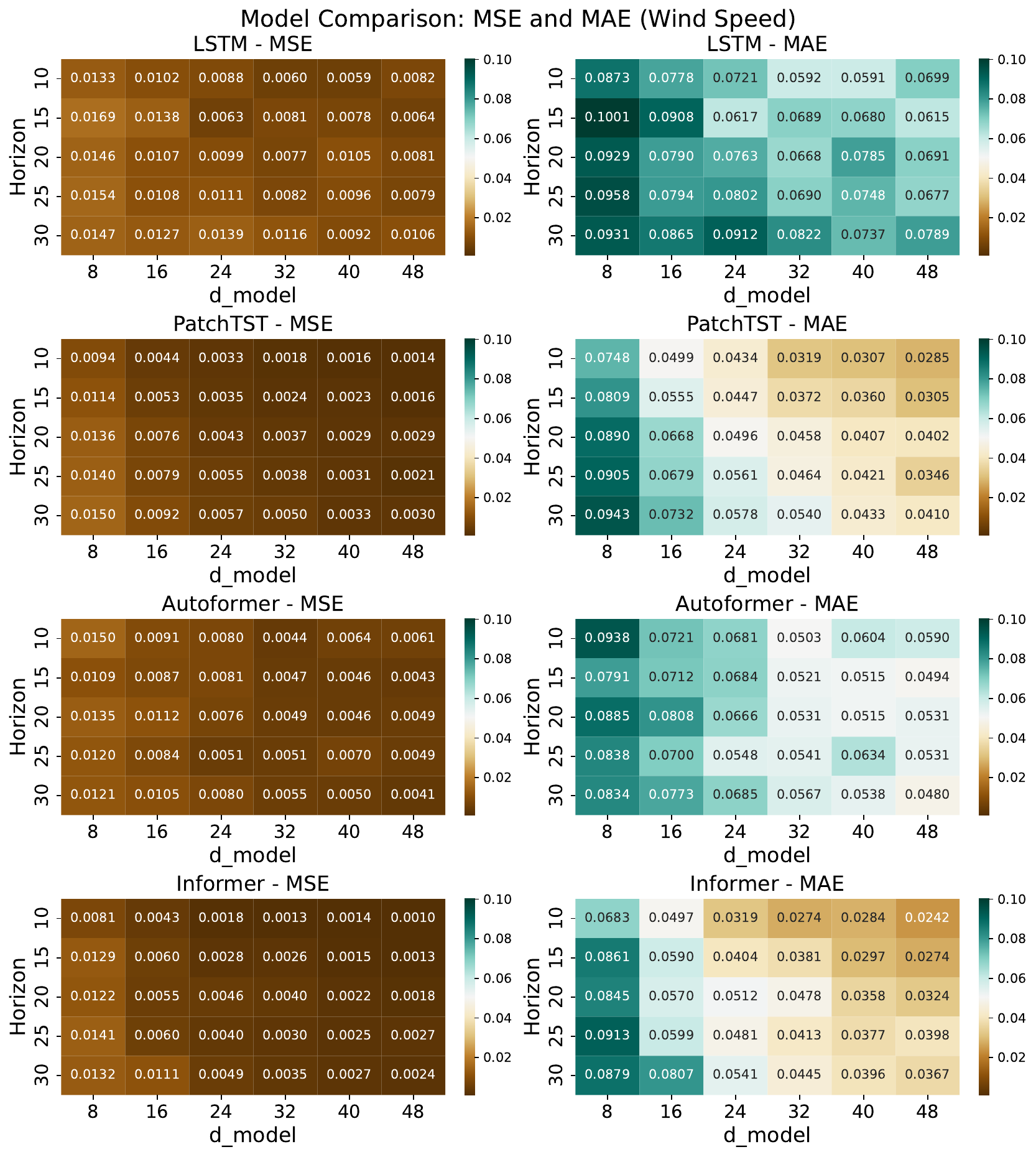}
    \caption{\dkoopformer versus LSTM baseline grid search for in CMIP6 dataset wind speed forecasting in scenario 2.}
    \label{fig:wind_speed_fixed_patch}
\end{figure*}

Figure~\ref{fig:wind_speed_fixed_patch} presents a comprehensive comparison of forecasting accuracy for wind speed using LSTM and three \dkoopformer~models (\patchtst, \autoformer, and \informer), with the hidden dimension ($d_{\text{model}}$) systematically varied to ensure fair capacity scaling across all architectures. The results demonstrate that all Koopman-augmented transformer models significantly outperform the LSTM baseline in both MSE and MAE across nearly all grid points, and this advantage becomes even more pronounced as $d_{\text{model}}$ and the forecast horizon increase. While increasing the LSTM's hidden size does improve its performance, it remains consistently less accurate than the transformer-based models, particularly for longer forecast horizons. Among the \dkoopformer~variants, \patchtst~and \informer~achieve the lowest error values, highlighting the benefit of patch-based and sparse attention mechanisms in capturing complex wind speed dynamics. \autoformer~also shows strong performance, typically surpassing LSTM but generally trailing \patchtst~and \informer. These findings highlight the superior scalability, accuracy, and robustness of Koopman-augmented transformer architectures for high-dimensional, multi-step wind speed forecasting, and demonstrate that, while LSTM benefits from increased capacity, it remains fundamentally limited by its sequential nature.

\begin{figure}
    \centering
    \includegraphics[width=0.99\linewidth]{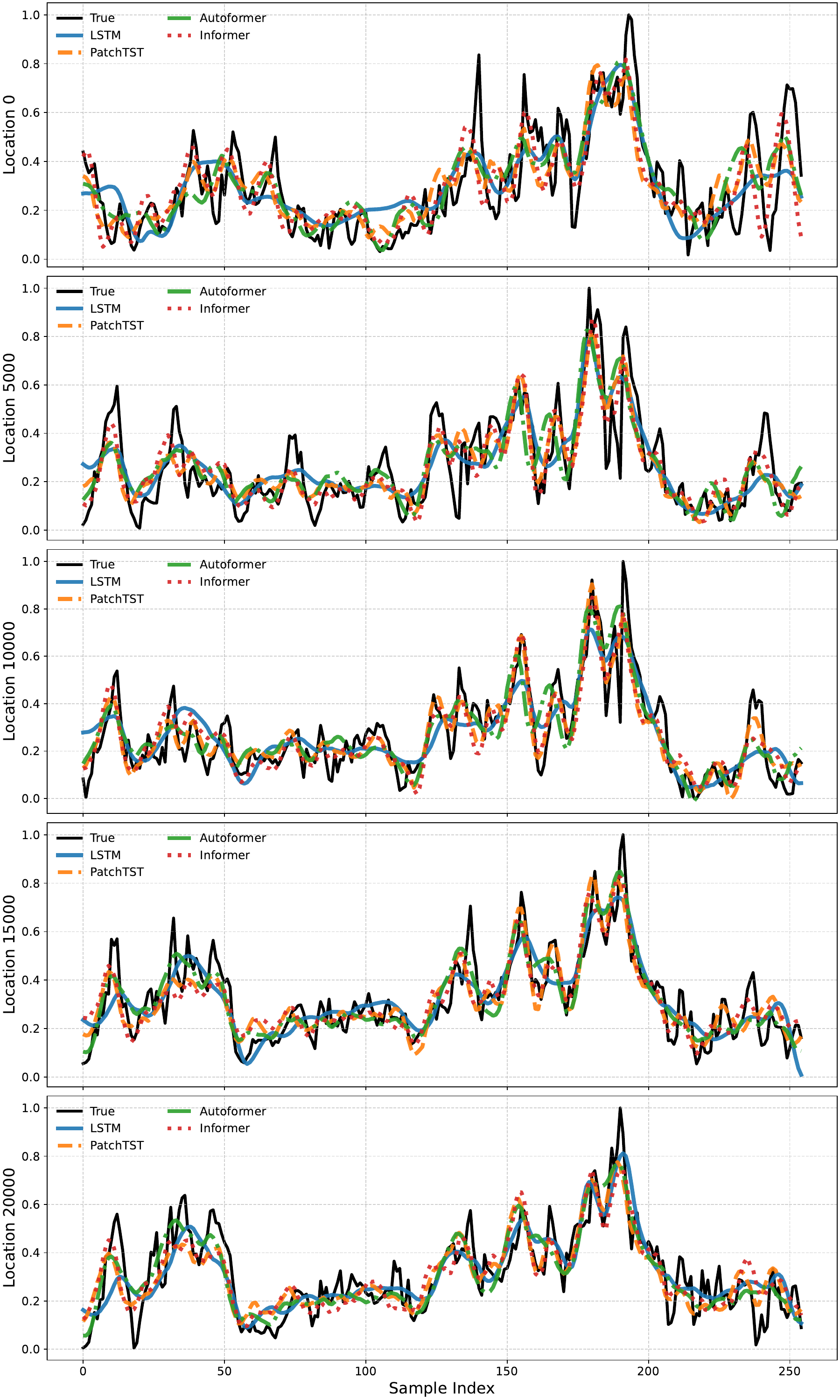}
    \caption{Comparison of \dkoopformer~with LSTM approach in wind speed for CMIP6 dataset forecasting with $p=120$, $H=25$ and $d_{\text{model}}=16$.}
    \label{fig:wind_speed_fixed_patch_1setting}
\end{figure}

Figure~\ref{fig:wind_speed_fixed_patch_1setting} illustrates representative forecast trajectories for five spatial locations using a patch length of $p=120$, a forecast horizon $H=25$, and $d_{\text{model}}=16$. Across all locations, the \dkoopformer~models closely track the true wind speed profiles and consistently outperform the LSTM, particularly in capturing rapid fluctuations and sharp peaks in the time series.

\subsubsection{Scenario 3}
For surface pressure forecasting, we employ a standardized experimental setup using four neural architectures: Koopman-augmented \patchtst, \autoformer, \informer, and a baseline LSTM. All models receive input sequences of fixed length ($P$ = 120) comprising up to five spatial channels, and are tasked with predicting multi-step future values across a range of forecast horizons ($H \in \{10, 15, 20, 25, 30\}$). For transformer-based architectures, the core latent dimension is set to \texttt{d\_model} = 48, with three encoder layers and four attention heads per layer. Patch-based models (\patchtst~ and \informer~) utilize variable patch lengths ($p \in \{80, 90, 100, 110, 120, 130\}$), where each patch is projected into the model's hidden space prior to temporal encoding. The \autoformer~leverages a moving average window of length $p$ for trend extraction. The LSTM baseline uses two layers with hidden dimension $64$, and all models use a linear decoder to map latent states to the multi-step output.

\begin{figure*}
    \centering
    \includegraphics[width=0.70\linewidth]{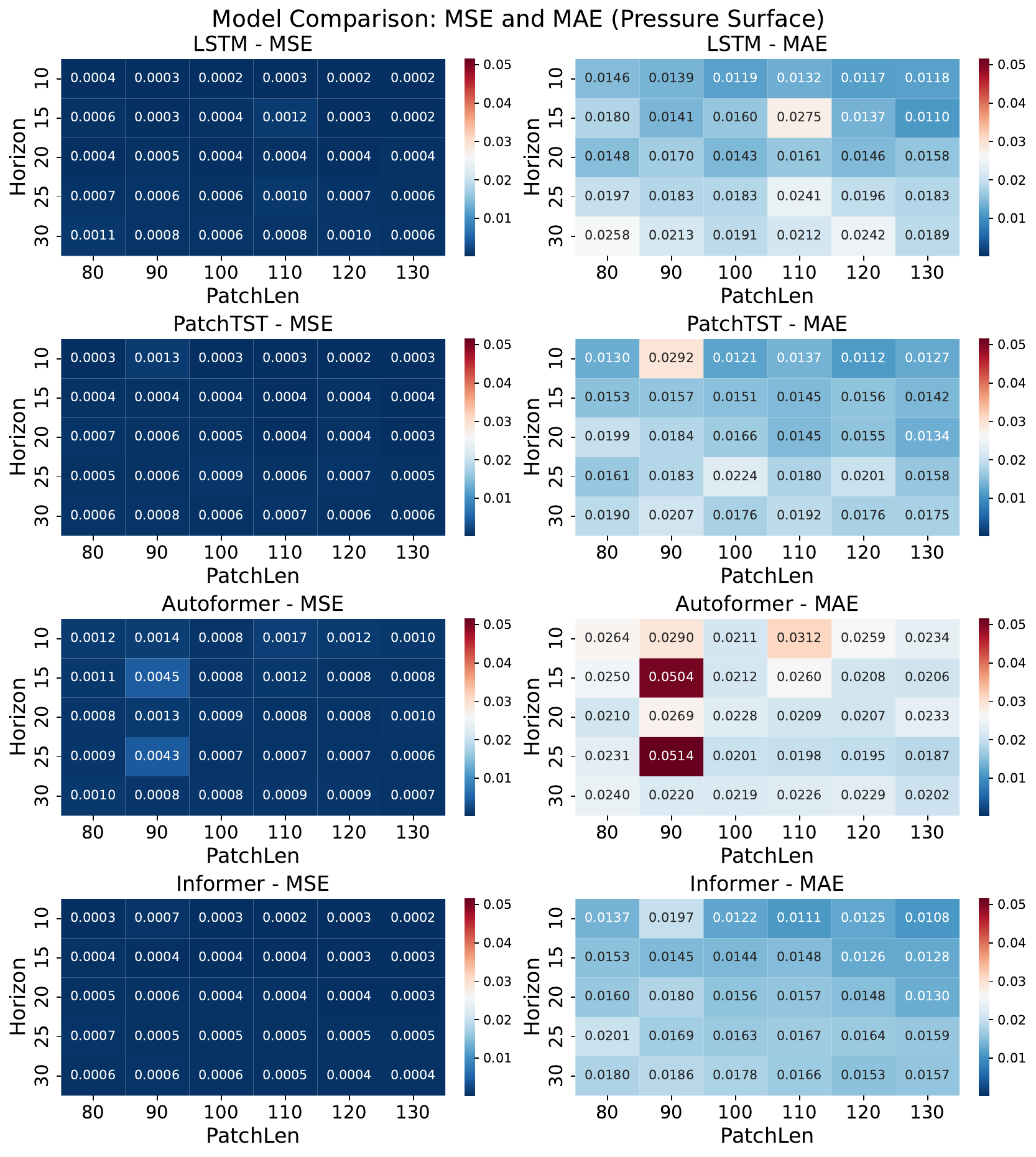}
    \caption{\dkoopformer~versus LSTM baseline grid search for pressure surface in CMIP6 dataset forecasting in scenario 3.}
    \label{fig:pressure_surface_fixed_dmodel}
\end{figure*}

\begin{figure}
    \centering
    \includegraphics[width=0.99\linewidth]{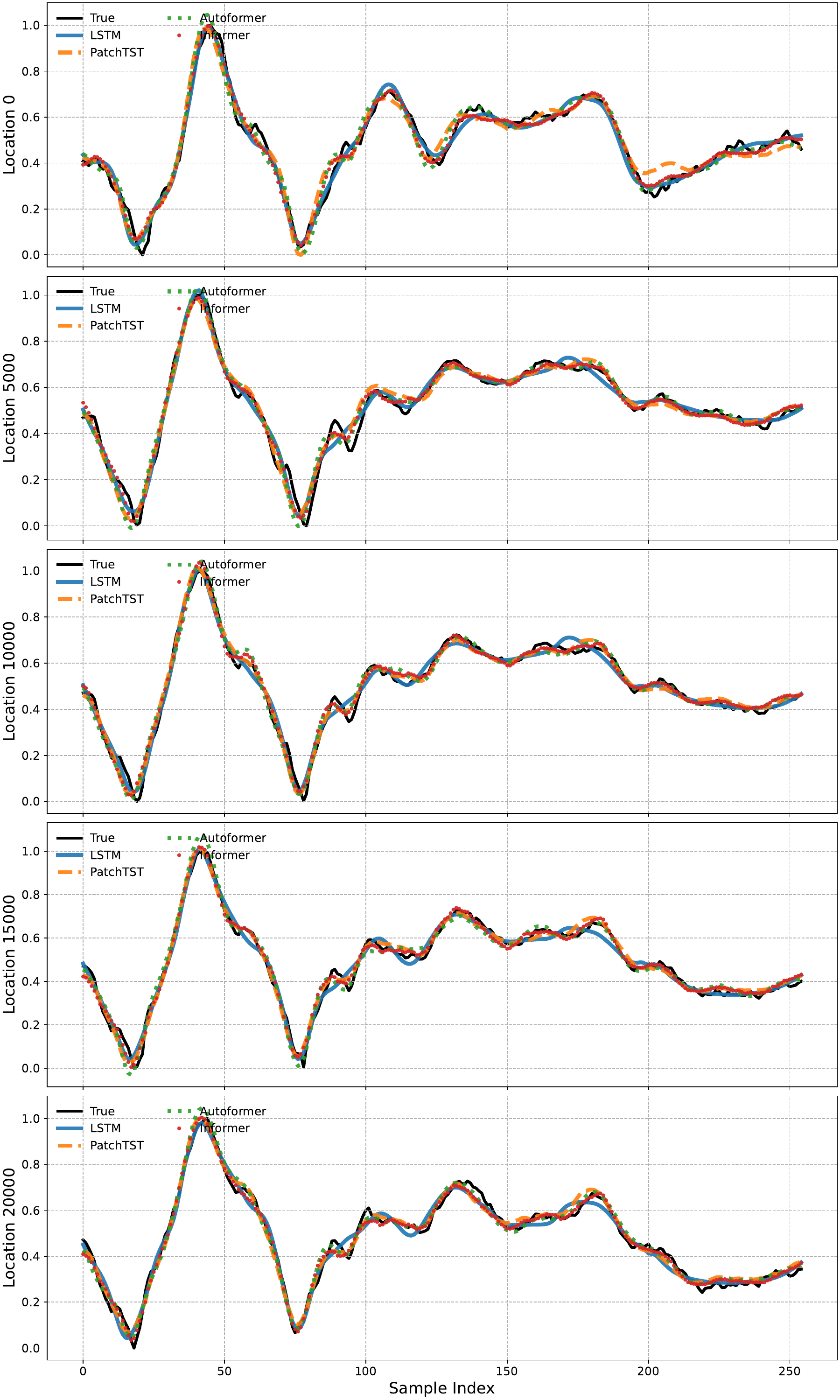}
    \caption{Comparison of \dkoopformer~pressure surface in CMIP6 dataset forecasting with $p=120$, $H=25$ and $d_{\text{model}}=48$, the LSTM hidden size is $64$.}
    \label{fig:pressure_surface_fixed_dmodel_1setting}
\end{figure}

The comparative results for surface pressure forecasting, shown in Fig. \ref{fig:pressure_surface_fixed_dmodel}, indicate that all models achieve remarkably low error rates (MSE and MAE) across the evaluated range of patch lengths and forecast horizons. Both the LSTM and \dkoopformer~models (\patchtst, \autoformer, \informer) demonstrate strong predictive accuracy, with MSE values often on the order of $10^{-4}$ to $10^{-3}$ and MAE consistently below $0.03$. Among the transformer variants, \patchtst~and \informer~deliver the lowest and most stable errors, maintaining performance across all patch length and horizon settings, and occasionally outperforming the LSTM, particularly for longer horizons and larger patch lengths. The \autoformer~model also shows competitive results but displays slightly higher MAE in some configurations, possibly due to its decomposition strategy being less advantageous for the smoother, less volatile pressure signals. Notably, the LSTM achieves comparable or even superior performance at shorter horizons and smaller patch sizes, highlighting its robustness on relatively stationary time series. Overall, these findings confirm that for surface pressure—where temporal dynamics are less abrupt and more predictable—both recurrent and transformer-based models can achieve excellent forecasting accuracy, with transformer models offering a marginal advantage in scenarios requiring longer-term prediction or increased model flexibility. Fig. \ref{fig:pressure_surface_fixed_dmodel_1setting} confirms that all the models have comparable performance.

\subsubsection{Scenario 4}
In the fourth scenario, we conducted a comprehensive grid search over major hyperparameters for all model variants (LSTM, and \dkoopformer~models). The primary parameters varied were the model dimension, with values $d_{\text{model}}= \{8, 16, 24, 32, 40, 48\}$, and the forecast horizon, ranging over $H= \{10, 15, 20, 25, 30\}$ time steps. Both the input sequence (context) length and patch length were fixed at 120 to ensure consistency across all models. For the LSTM baseline, the hidden size was set equal to $d_{\text{model}}$, enabling a fair comparison of model capacity across all architectures. 

\begin{figure*}[!t]
    \centering
    \includegraphics[width=0.70\linewidth]{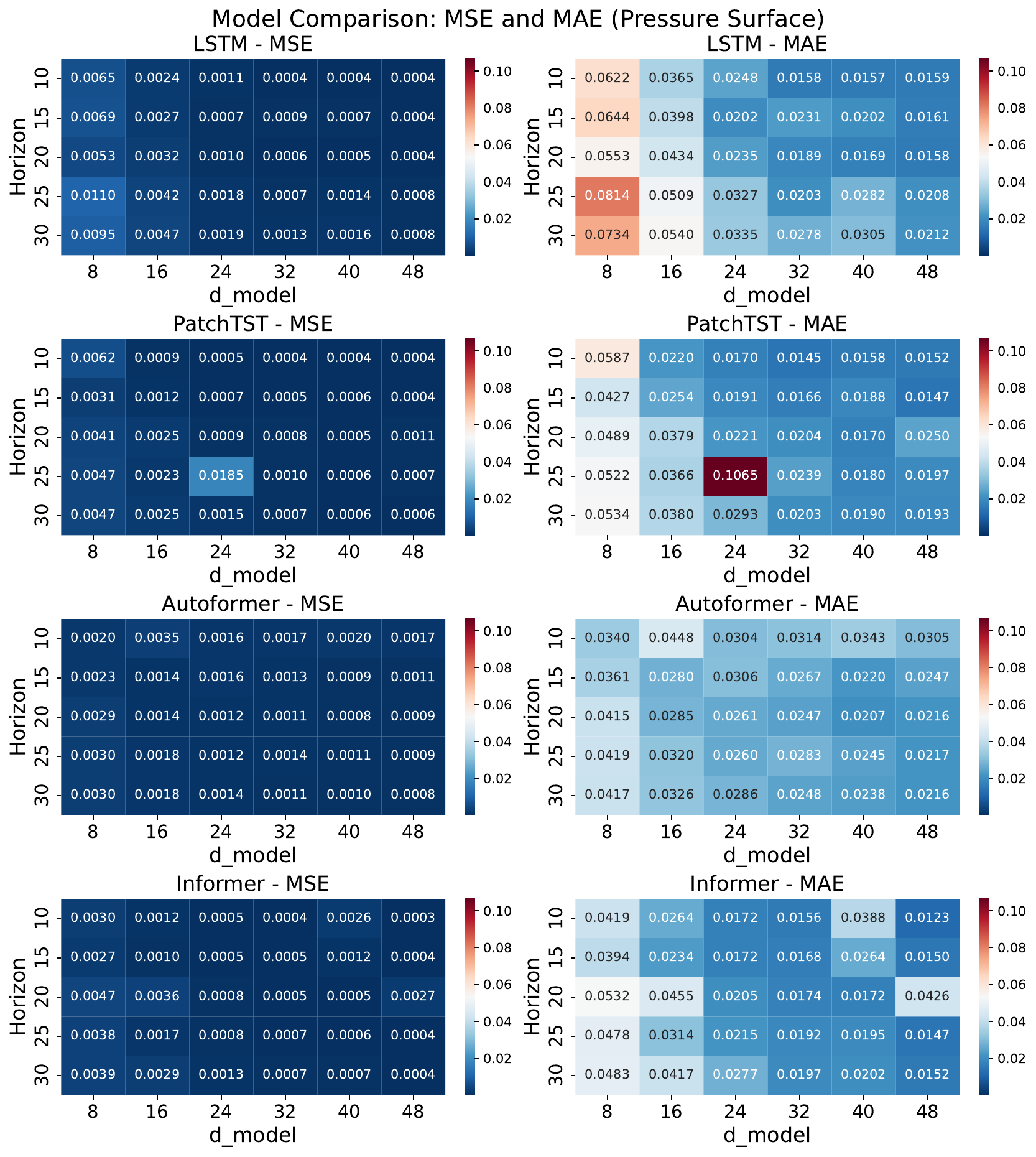}
    \caption{\dkoopformer~versus LSTM baseline grid search for CMIP6 wind speed forecasting in scenario 4.}
    \label{fig:pressure_surface_fixed_patch}
\end{figure*}

The heatmap results for the pressure surface forecasting task (Fig. \ref{fig:pressure_surface_fixed_patch}) reveal several key trends across all model families as both the model dimension ($d_{\mathrm{model}}$) and forecast horizon vary. All models demonstrate improved performance (lower MSE and MAE) as $d_{\mathrm{model}}$ increases, particularly at shorter horizons. The \dkoopformer~variants consistently outperform the LSTM baseline across nearly all hyperparameter settings, with noticeably lower MSE and MAE values at higher model widths and shorter to moderate horizons. \informer~and \autoformer~achieve the lowest overall errors, highlighting the benefit of transformer-based architectures with Koopman integration for capturing complex spatiotemporal dependencies in atmospheric pressure fields. The LSTM shows relatively stable but higher error, especially at larger horizons and lower $d_{\mathrm{model}}$. Outliers in MAE, such as at $(d_{\mathrm{model}}=24,\,\mathrm{horizon}=25)$ for \patchtst, may be attributed to local convergence issues or data variance at that configuration. Overall, these results confirm the effectiveness of \dkoopformer~models in achieving robust and accurate multi-step pressure forecasting, especially when sufficient model capacity is available.

\begin{figure}[!t]
    \centering
    \includegraphics[width=0.99\linewidth]{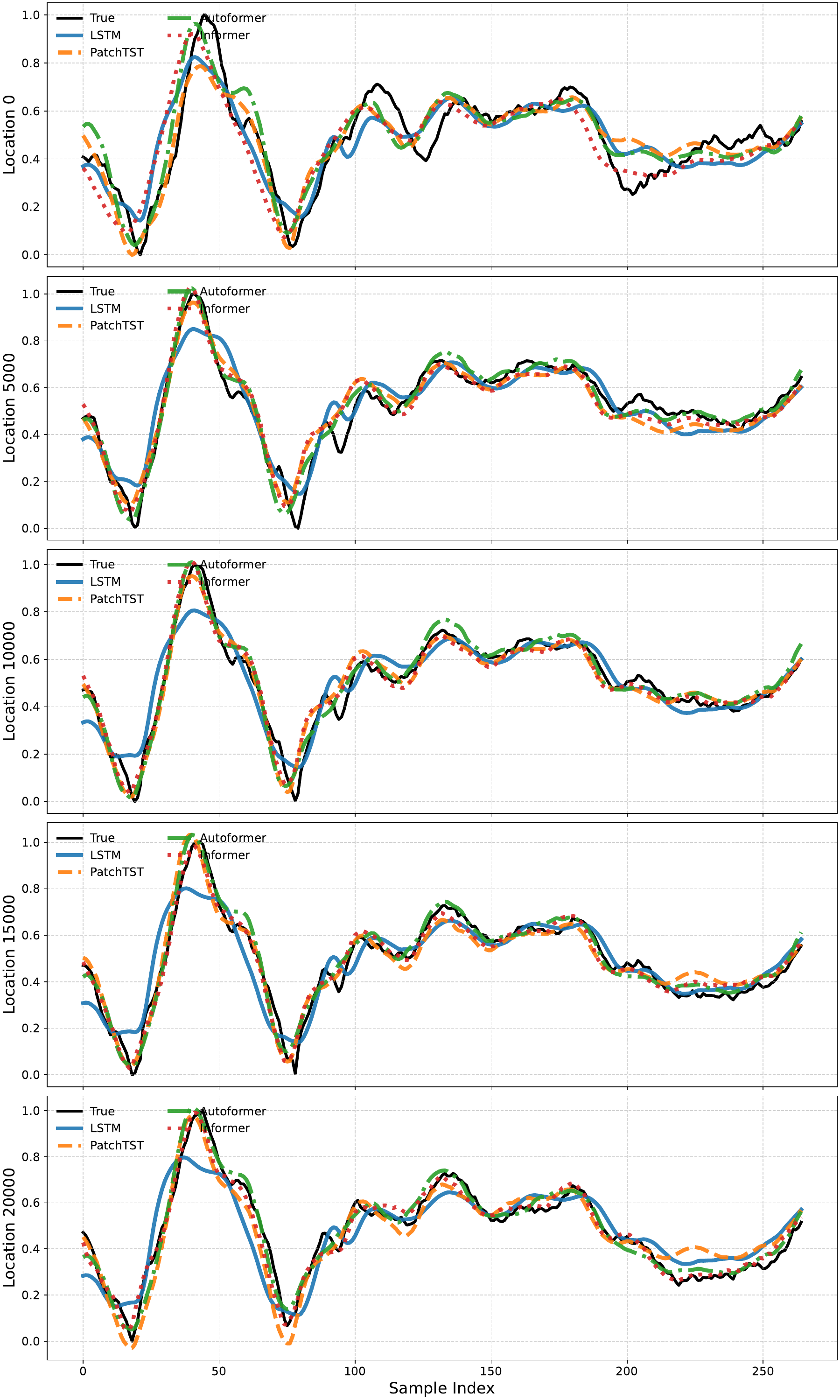}
    \caption{
    Comparison of \dkoopformer~pressure surface forecasting in CMIP6 dataset with $p=120$, $H=15$ and $d_{\text{model}}=8$; LSTM hidden size is $8$.}
    \label{fig:pressure_surface_fixed_patch_1setting}
\end{figure}

Fig. \ref{fig:pressure_surface_fixed_patch_1setting} depicts the predicted pressure surface time series at five representative spatial locations for patch length $p= 120$, $d_{\mathrm{model}} = 8$, and forecast horizon $H = 15$. All \dkoopformer~models track the ground truth signal closely, and generally provide more accurate and less noisy forecasts compared to the LSTM baseline.

\subsubsection{Scenario 5}
In this scenario we test the \dkoopformer~models on the ERA5 reanalysis wind speed dataset over Germany. We choose $10$ different locations over the map and feed into our framework. The LSTM comprises a two-layer LSTM with $48$ hidden units per layer and a linear projection mapping the final hidden state to the output. For the \dkoopformer~models, each backbone consists of 3 Transformer encoder layers, each with four attention heads and a latent dimension ($d_{\mathrm{model}}$) of $48$. Feedforward (hidden) layers use a width of 96. Sinusoidal positional encodings are applied throughout. Patch-based variants (\patchtst~ and \informer) use a one-dimensional convolution to embed the sequence into non-overlapping patches of length $p$ (with stride $p$); for \autoformer, the patch length sets the moving average kernel in its decomposition module.

\dkoopformer~variants append a strictly stable Koopman operator, parametrized as a learnable linear map with spectral radius $\rho_{\max}=0.99$, acting on the latent space before the output projection. All models are trained for a user-specified number of epochs $4000$ using the Adam optimizer (learning rate $3 \times 10^{-4}$). \dkoopformer~variants use a Lyapunov regularization term (weight $\lambda= 0.1$) during training.

\begin{figure}[h]
    \centering
    \includegraphics[width=0.75\linewidth]{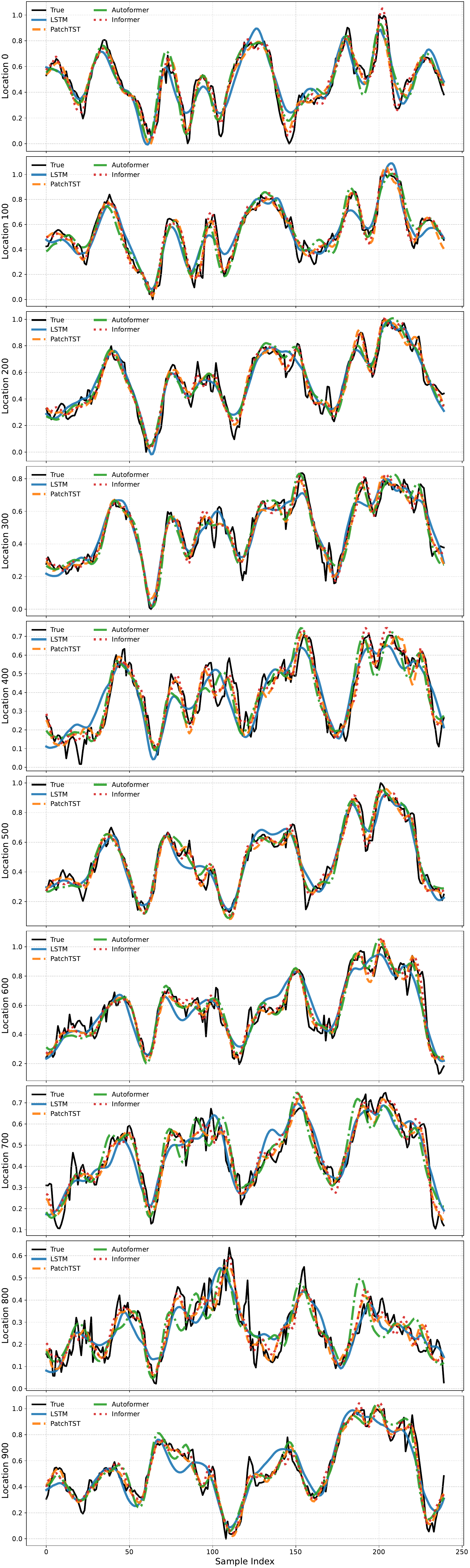}
    \caption{Comparison of \dkoopformer~models with LSTM for wind speed forecasting at 10 representative locations in Germany with ERA5 dataset in scenario 5.}
    \label{fig:wind_eval_era5}
\end{figure}

The multi-location wind speed forecasts presented in Fig.~\ref{fig:wind_eval_era5} demonstrate the comparative performance of the LSTM and \dkoopformer~models across 10 spatial sites. Visually, all \dkoopformer~variants are able to closely track the ground truth signal and effectively capture both the amplitude and timing of the primary wind fluctuations, with their predicted trajectories generally overlapping the reference signal throughout the sample window. The LSTM baseline also exhibits strong trend-following capability, but tends to produce slightly smoother predictions that sometimes lag during rapid transitions or sharp peaks, particularly at locations with more pronounced variability. Among the \dkoopformer~architectures, \patchtst~and \informer~typically yield the most faithful short-term tracking and local variability, while \autoformer~occasionally exhibits more smoothed outputs and, at certain sites, larger deviations near abrupt signal changes. Overall, the Koopman-enhanced transformers systematically match or surpass the LSTM baseline, especially in reproducing high-frequency fluctuations and transient behavior, illustrating the benefit of combining deep temporal modeling with latent linear operator propagation for spatiotemporal forecasting.

\subsubsection{Scenario 6}
In this scenario, the dataset is partitioned into training and testing splits with an 80\%/20\% ratio. All experiments use a multivariate wind speed dataset sourced from CMIP, consisting of five channels (spatial points), each with $P=2920$ samples corresponding to a single year of data at 3-hour intervals. To ensure fair comparison, identical data preprocessing and scaling are applied across all models. 

The primary hyperparameters include the number of encoder layers set to 3 for \dkoopformer-based models, the number of attention heads 4 per encoder layer, and the dimensionality of latent embeddings $d_{\text{model}}=48$. For temporal segmentation, the patch length (\texttt{patch\_len}) is varied within the set $p=\{40, 50, 60, 70, 80, 90, 100\}$, and forecast horizons (\texttt{horizon}) are swept over $H=\{2, 4, 6, 8, 10, 12, 14, 16\}$. Each experiment uses a fixed input sequence length ($\texttt{seq\_len} = 120$), and models are trained for up to $4000$ epochs using the Adam optimizer with a learning rate of $3 \times 10^{-4}$.

For LSTM, a two-layer architecture with 48 hidden units is used. The transformer-based backbones leverage positional encodings and, where applicable, patch embeddings or series decomposition. All Koopformer variants incorporate a strictly stable Koopman operator, regularized via a Lyapunov term with weight 0.1. Evaluation metrics—including mean squared error (MSE) and mean absolute error (MAE)—are reported for each model, patch length, and forecast horizon on both train and test sets.

\begin{figure*}[h]
    \centering
    \includegraphics[width=0.75\linewidth]{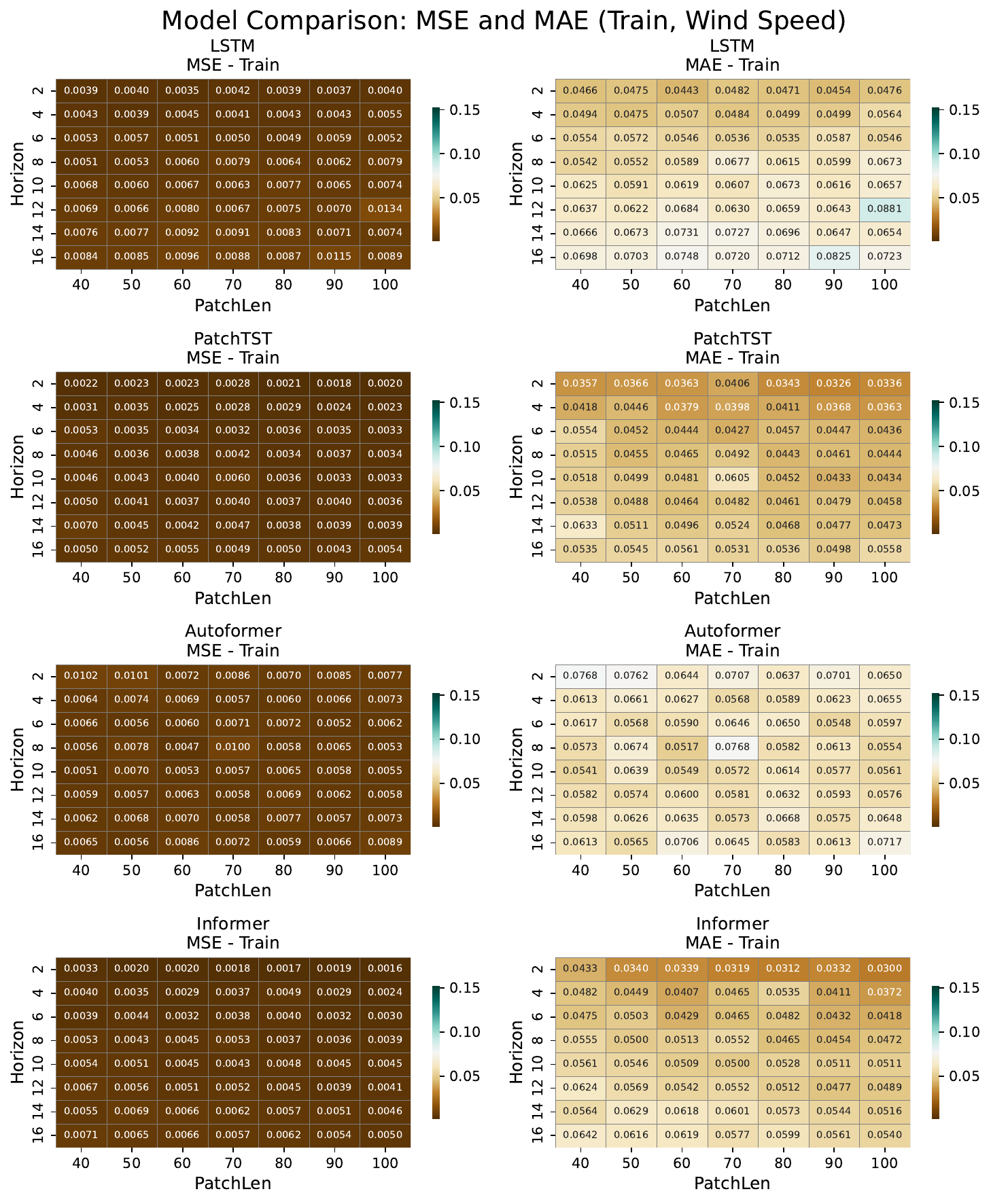}
    \caption{\dkoopformer~models versus LSTM baseline grid search for CMIP6 wind speed forecasting in scenario 6.}
    \label{fig:wind_cimip_train_80percent_full_year}
\end{figure*}

Figure~\ref{fig:wind_cimip_train_80percent_full_year} summarizes the training performance of all evaluated models on the five-channel CMIP wind speed dataset.


\dkoopformer-\patchtst~achieves consistently low training errors, particularly at moderate patch lengths and short to mid-range horizons, indicating effective learning of spatiotemporal dynamics.
\dkoopformer-\autoformer~shows slightly higher error magnitudes but remains stable across patch sizes, benefitting from its trend-seasonal decomposition. \dkoopformer-\informer~also exhibits competitive performance, particularly at smaller patch lengths, reflecting the efficacy of self-attention even without aggressive patching. In contrast, the LSTM baseline incurs higher training errors across nearly all configurations, particularly for longer forecast horizons, confirming the advantage of transformer-based and Koopman-regularized architectures. The overall trends suggest that Koopman operator constraints and patch-based temporal abstraction significantly enhance training fidelity and convergence stability in multivariate wind speed forecasting.

\begin{figure*}[h]
    \centering
    \includegraphics[width=0.75\linewidth]{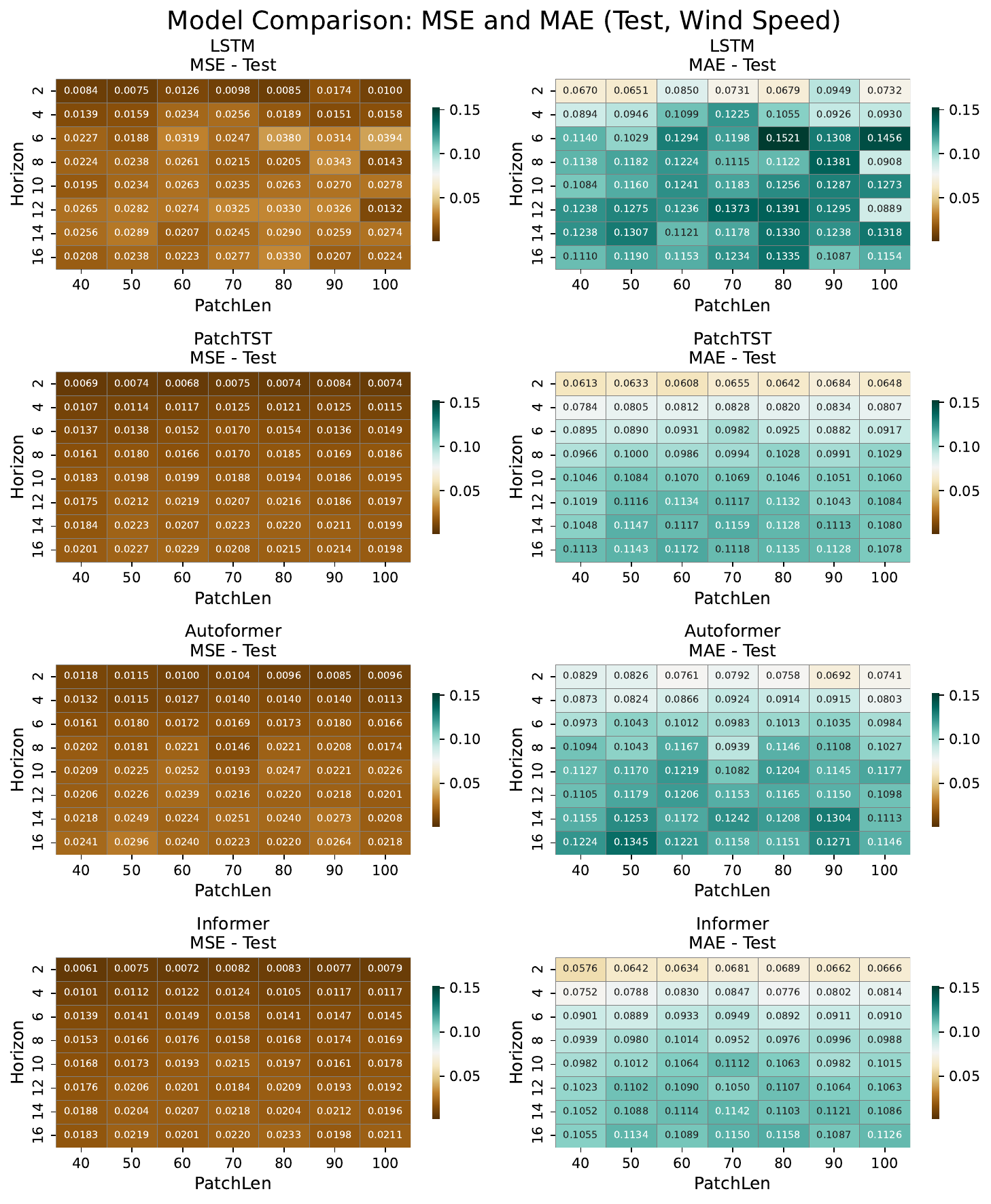}
    \caption{\dkoopformer~models versus LSTM baseline grid search in test dataset for CMIP6 wind speed forecasting in scenario 6.}
    \label{fig:wind_cimip_test_evaluation}
\end{figure*}

Figure~\ref{fig:wind_cimip_test_evaluation} illustrates the forecasting performance of all Koopformer variants and the LSTM baseline on the test set using the CMIP-derived five-channel wind speed dataset. Each heatmap reports the mean absolute error (MAE) or mean squared error (MSE) across a range of patch lengths $p$ (horizontal axis) and forecast horizons $H$ (vertical axis). The heatmaps reveal distinct patterns in predictive performance across models. \dkoopformer-\patchtst~consistently achieves lower test errors across most configurations, particularly for shorter horizons and moderate patch lengths. However, \autoformer~and \informer~demonstrate stable performance but are more sensitive to increasing forecast horizons. In contrast, the LSTM baseline shows higher errors, especially for longer horizons, indicating limited generalization capacity. The color gradients and annotated cell values in each heatmap allow for precise comparison across configurations. These results confirm the advantages of Koopman operator integration with transformer backbones, particularly when combined with patch-based temporal embeddings. Notably, optimal performance is achieved when patch lengths are appropriately aligned with the underlying temporal scale of variability in the input data.

\begin{figure}[h]
    \centering
    \includegraphics[width=0.75\linewidth]{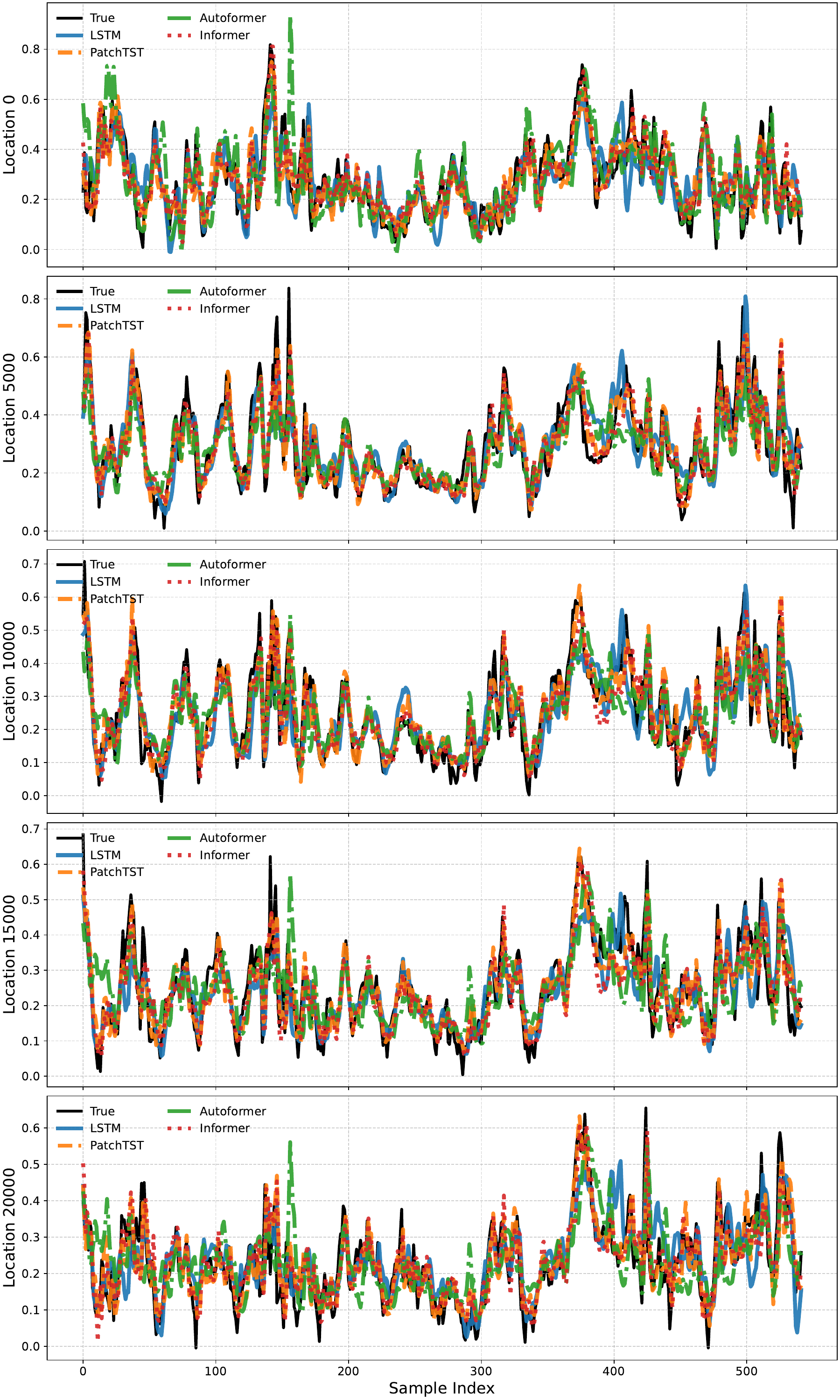}
    \caption{Comparison of \dkoopformer~on the test dataset for CMIP6 wind speed forecasting with $p = 40$, $H = 8$ and $d_{\text{model}}=48$; the LSTM
hidden size is $48$.}
    \label{fig:wind_cimip_test_evaluation_test_patch40_h8}
\end{figure}

Fig.~\ref{fig:wind_cimip_test_evaluation_test_patch40_h8} presents the forecasting performance of all evaluated models on the test set for a fixed patch length of $p=40$ and forecast horizon of $H=8$ time steps, using five spatial channels extracted from the CMIP wind speed dataset. Among the tested models, \dkoopformer-\patchtst~demonstrates the most accurate forecasts, closely tracking the ground truth across all spatial locations. \dkoopformer-\informer~performs competitively, maintaining good temporal alignment, though with occasional fluctuations in amplitude. In contrast, \dkoopformer-\autoformer~consistently exhibits lower performance, often underestimating the signal and failing to capture the full dynamics, especially in more volatile regions. The LSTM baseline performs slightly better than \autoformer~in some cases but generally shows delayed responses and smoothed predictions. These results highlight that \dkoopformer~models based on \patchtst~and \informer~architectures benefit more from Koopman operator integration and patch-based attention mechanisms, whereas the \autoformer~backbone—despite its trend-seasonality decomposition—struggles to generalize in this multivariate wind forecasting task under moderate horizon and patch configurations.

\subsubsection{Scenario 7}
\begin{figure*}
     \centering
    \includegraphics[width=0.75\linewidth]{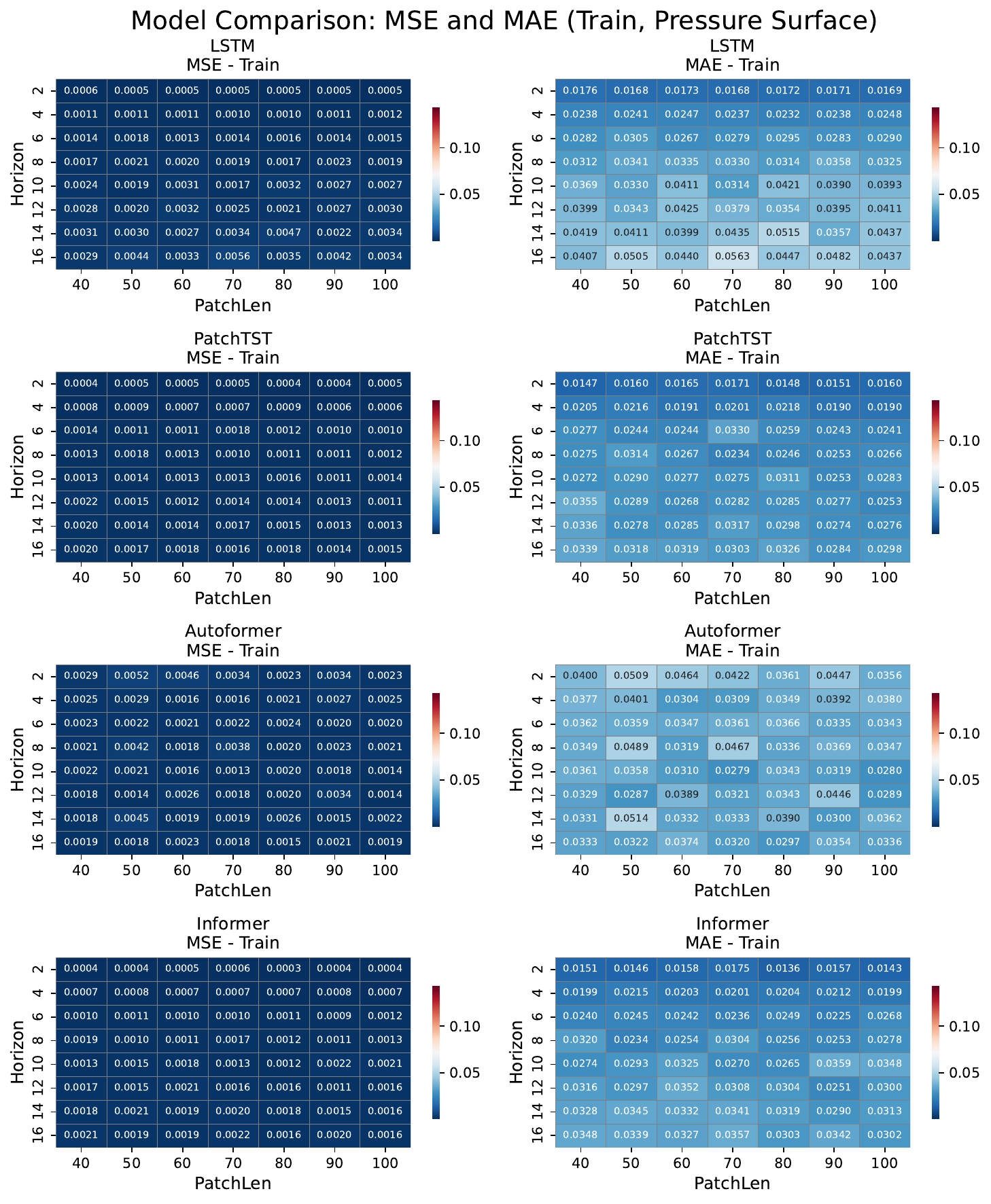}
    \caption{\dkoopformer~models versus LSTM baseline grid search in train dataset for CMIP6 pressure surface forecasting in scenario 7.}
    \label{fig:pressure_cimip_train_80percent_full_year}
\end{figure*}

Fig~\ref{fig:pressure_cimip_train_80percent_full_year} reports the training set performance of all models on the surface pressure forecasting task, evaluated across various patch lengths ($p \in \{40, 50, 60, 70, 80, 90, 100\}$) and forecast horizons ($H \in \{2, 4, 6, 8, 10, 12, 14, 16\}$). The heatmaps display the mean squared error (MSE) and mean absolute error (MAE) for each configuration. Among the evaluated models, \patchtst~consistently achieves the lowest training errors, especially for short-to-medium horizons, indicating strong learning capacity and low bias. \dkoopformer-\informer~also performs competitively across many configurations, particularly with patch lengths around $p=60–80$. In contrast, \dkoopformer-\autoformer~shows more variability in performance and often lags behind other transformer-based variants, particularly for longer horizons. The LSTM baseline exhibits higher training errors overall compare to \informer~and \patchtst~models, reflecting its limited ability to capture complex temporal dependencies in the pressure field data. These results confirm the advantage of Koopman-regularized transformer architectures, particularly those that incorporate patch-wise tokenization and direct attention mechanisms, in learning smooth and spatially coherent pressure dynamics during training.

\begin{figure*}
    \centering
    \includegraphics[width=0.75\linewidth]{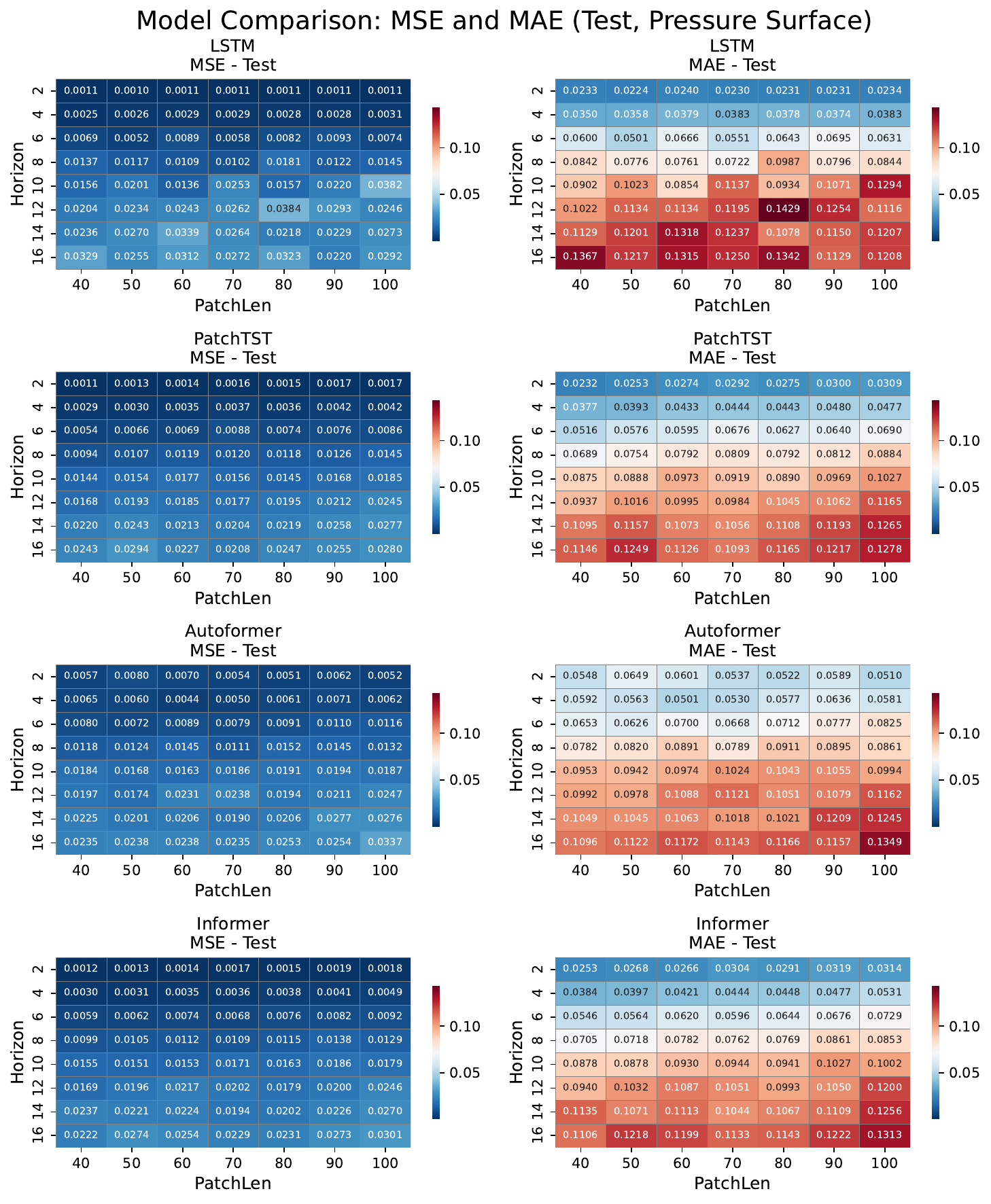}
    \caption{\dkoopformer~models versus LSTM baseline grid search in test dataset for CMIP6 pressure surface forecasting in scenario 7.}
    \label{fig:pressure_cimip_test_evaluation}
\end{figure*}

Fig.~\ref{fig:pressure_cimip_test_evaluation} presents the test set forecasting performance of all evaluated models on the surface pressure dataset. Among all models, \dkoopformer-\patchtst~achieves the lowest error values across most configurations, particularly at short and medium forecast horizons, indicating superior generalization from training to unseen test sequences. \informer~follows with relatively competitive performance but shows slightly elevated errors as the forecast horizon increases. \autoformer~consistently underperforms compared to the other \dkoopformer~variants, with error magnitudes increasing sharply for longer horizons and larger patch lengths, suggesting instability in trend decomposition when extrapolating beyond training data. The LSTM baseline demonstrates the highest test errors overall, confirming its limited capacity to capture complex spatiotemporal dependencies inherent in pressure surface dynamics. These findings further reinforce the benefit of integrating Koopman constraints with attention-based sequence modeling, particularly when coupled with patch-based input representations optimized for multi-horizon forecasting.

\begin{figure}
    \centering
    \includegraphics[width=0.75\linewidth]{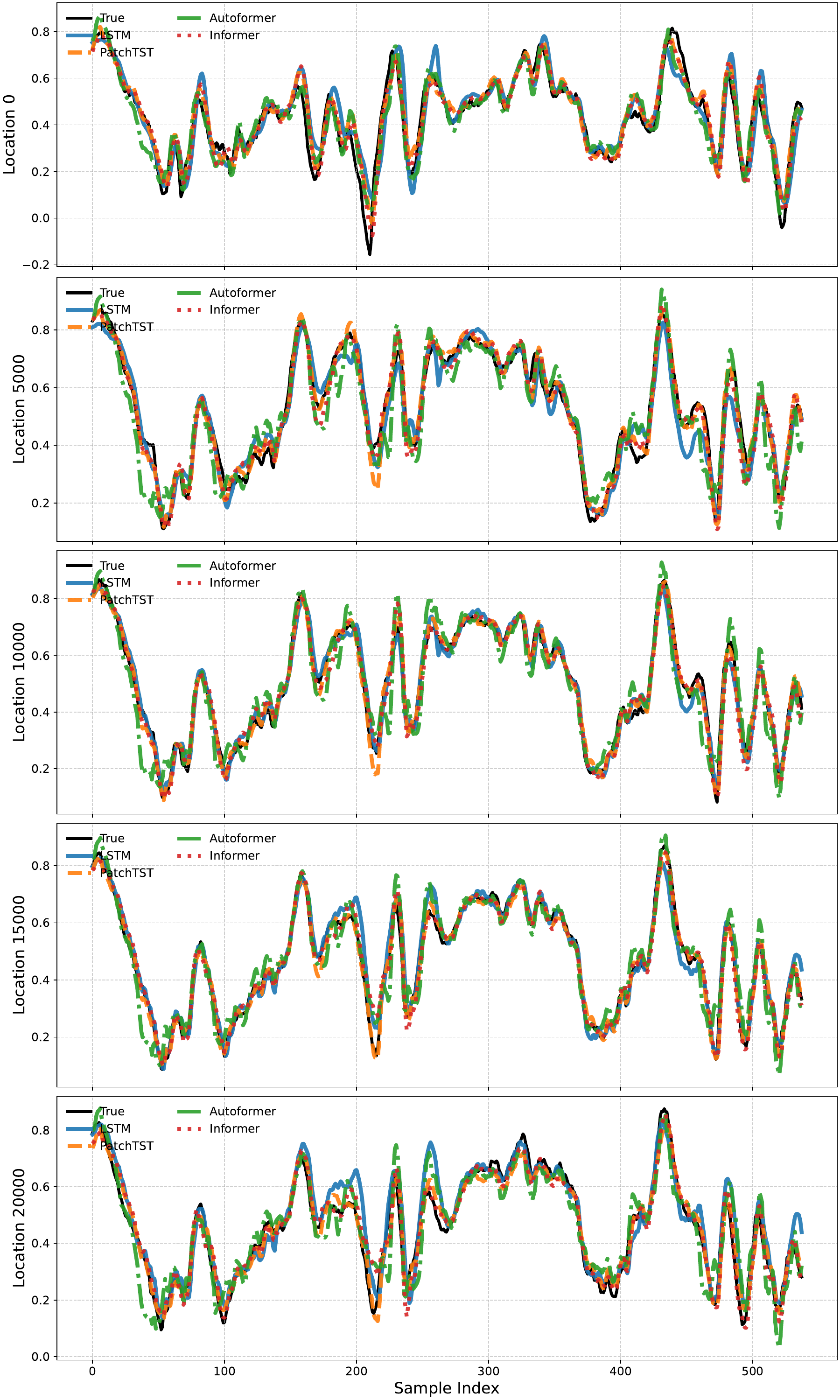}
    \caption{Comparison of \dkoopformer~on the test dataset for CMIP6 pressure surface forecasting with $p = 40$, $H = 8$ and $d_{\text{model}}=48$; the LSTM hidden size is $48$.}
    \label{fig:pressure_cimip_test_evaluationtest_patch40_h8_d48}
\end{figure}

Figure~\ref{fig:pressure_cimip_test_evaluationtest_patch40_h8_d48} displays the test set predictions of all models on the surface pressure dataset for a forecast horizon of 8 time steps and a patch length of 40. The experiment was conducted on five selected spatial indices (\texttt{0}, \texttt{5000}, \texttt{10000}, \texttt{15000}, \texttt{20000}), with a shared embedding dimension of $d=48$. For each spatial location, the predicted sequences from LSTM, \patchtst, \autoformer, and \informer~are plotted against the ground truth. \patchtst~provides the most faithful reconstruction of the true signal across all spatial features, with predictions that tightly follow temporal variations. \informer~also performs well, showing relatively accurate tracking though with minor deviations near peak values. \autoformer, however, exhibits larger deviations, with smoothed or lagging outputs, particularly in high-amplitude regions. The LSTM baseline shows consistent underperformance, struggling to capture rapid variations and often underestimating the amplitude of dynamic fluctuations. These results reaffirm that patch-based transformers with Koopman constraints, especially Koopformer-PatchTST, are well-suited for multivariate pressure forecasting tasks with medium-range horizons.

\subsection{Application of \dkoopformer~on financial dataset}
We utilize a comprehensive cryptocurrency market dataset, comprising 27{,}585 records of daily transaction data for the top 10 cryptocurrencies traded on the Binance Exchange from September 1, 2018 to September 5, 2022. The dataset includes key market indicators such as opening, closing, high, and low prices, trading volume, and market capitalization, providing a rich temporal and cross-sectional representation of cryptocurrency market dynamics \footnote{\url{https://github.com/Chisomnwa/Cryptocurrency-Data-Analysis}}. 

\subsubsection{Scenario 8}
For our experiments, we used the \texttt{Crypto Coin} dataset and specifically preprocessed the \texttt{ClosePrice} series by splitting it into two equal-length segments of 2{,}900 entries each. This resulted in a two-channel array of shape $(2900,\, 2)$, which enabled the \dkoopformer~model to learn from parallel streams of historical price data in a multi-channel forecasting setup. Leveraging this dataset, we train and evaluate the \dkoopformer~algorithm. The dataset is partitioned into training and testing splits with an 80\%/20\% ratio.

The primary hyperparameters for the \dkoopformer-based models include an encoder comprising 3 layers, each with 4 attention heads, and a latent embedding dimensionality of $d_{\text{model}} = 48$. Temporal segmentation is achieved by varying the patch length within the set $p = \{40,\, 50,\, 60,\, 70,\, 80,\, 90,\, 100\}$, while the forecast horizon (\texttt{horizon}) is swept over $H = \{2,\, 4,\, 6,\, 8,\, 10,\, 12,\, 14,\, 16\}$. All experiments use a fixed input sequence length, and models are trained for up to $4{,}000$ epochs using the Adam optimizer with a learning rate of $3 \times 10^{-4}$.

For the LSTM baseline, we employ a two-layer architecture with 48 hidden units per layer. Transformer-based models utilize positional encodings and, where relevant, patch embeddings or series decomposition techniques. All \dkoopformer~variants enforce a strictly stable Koopman operator, regularized by a Lyapunov penalty with a weight of $\lambda= 0.1$. Model performance is evaluated using mean squared error (MSE) and mean absolute error (MAE), with results reported for each combination of model type, patch length, and forecast horizon on both training and test sets.

The diversity and volatility present in the data---especially across different coins and market conditions---enables a rigorous assessment of the model's ability to capture complex, nonlinear patterns in financial time series. The insights derived from this real-world dataset helped demonstrate the effectiveness and generalization capability of \dkoopformer~in forecasting and analyzing cryptocurrency price trends under various market regimes.

\begin{figure*}
    \centering
    \includegraphics[width=0.75\linewidth]{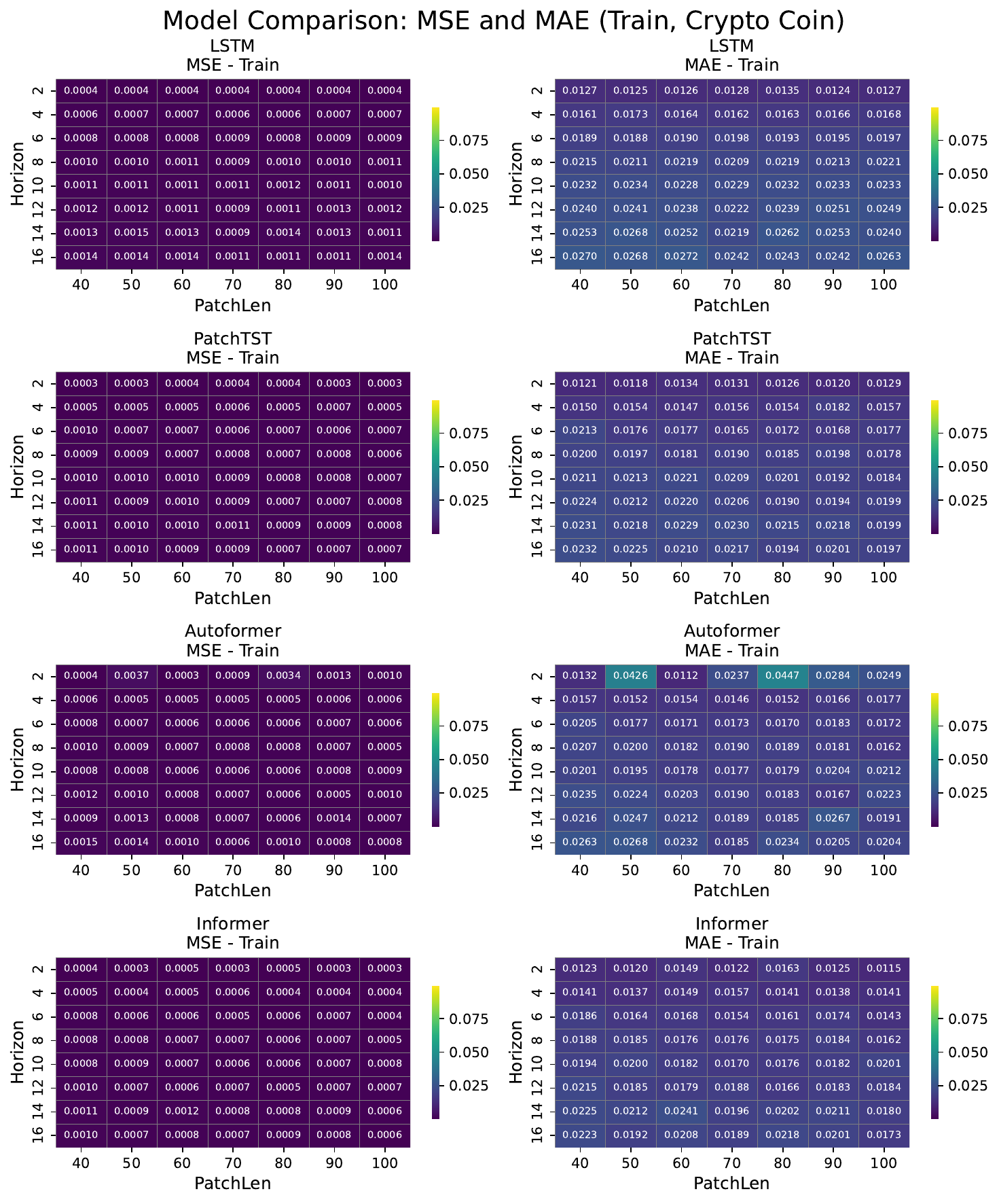}
    \caption{\dkoopformer~models versus LSTM baseline grid search in train dataset for financial time series forecasting in scenario 8.}
    \label{fig:Crypto_pred_train_grid}
\end{figure*}

The training heatmaps \ref{fig:Crypto_pred_train_grid} show that all models achieve low error values across the explored patch lengths and horizons, indicating successful overfitting or memorization of the training data. LSTM consistently reports the lowest MSE values across the board, particularly in shorter horizons (e.g., $H \leq 6$), where its simplicity aligns well with shorter-term temporal dependencies. \dkoopformer-based \patchtst~models maintain competitive MSE and MAE values, particularly at longer patch lengths (e.g., $p \geq 80$), suggesting they benefit from learning richer temporal structures. \autoformer~and \informer~also show strong performance, but \autoformer~displays slightly more instability in MAE across certain settings. Interestingly, \patchtst~and \informer~variants exhibit robust and stable error surfaces, demonstrating their capacity to extract and encode latent structures over a wide horizon and patch range. Overall, these results confirm that all Transformer variants, especially \dkoopformer-based ones, effectively leverage their inductive biases during training on cryptocurrency dynamics.

\begin{figure*}
    \centering
    \includegraphics[width=0.75\linewidth]{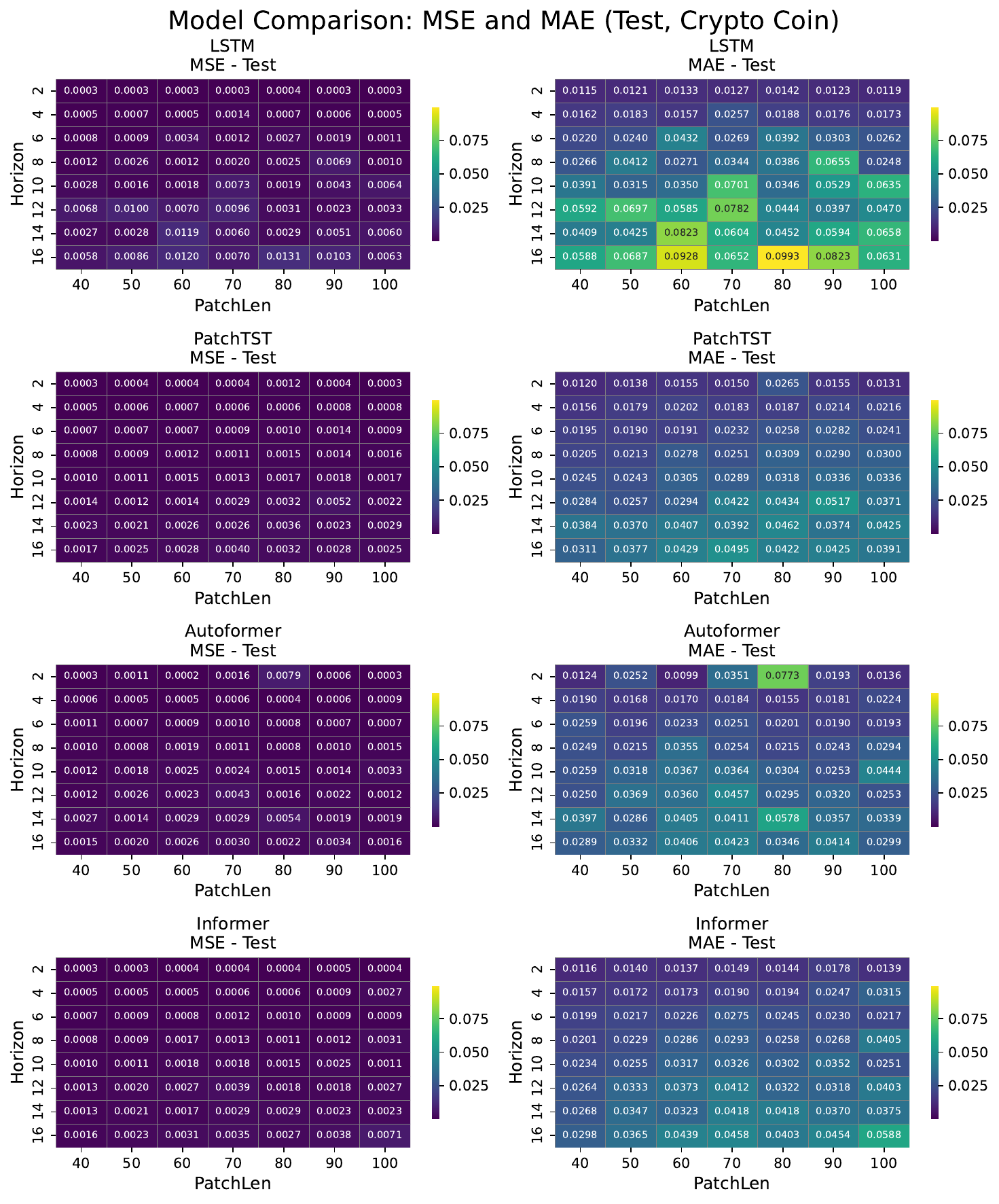}
    \caption{\dkoopformer~models versus LSTM baseline grid search in test dataset for financial time series forecasting in scenario 8.}
    \label{fig:Crypto_pred_test_grid}
\end{figure*}

On the test set shown in Fig. \ref{fig:Crypto_pred_test_grid}, the differences in generalization performance become more pronounced. LSTM, while effective in training, suffers a significant degradation in MAE as both the patch length and forecast horizon increase, reflecting its limited capacity to generalize beyond short-term patterns. \dkoopformer~variants, especially \patchtst, display strong generalization across a wide range of configurations, maintaining low MSE and MAE, particularly at higher patch lengths ($p \geq 70$) and longer horizons. In contrast, \autoformer~and \informer~models exhibit higher test errors, with \autoformer~occasionally peaking in MAE (e.g., $H = 2$, $p = 80$), suggesting potential overfitting or instability under certain configurations. The \dkoopformer’s ability to maintain low test errors across different regimes highlights its strength in capturing the intrinsic Koopman dynamics underlying complex, volatile price trends. This robustness is critical for deployment in financial time series forecasting, where adaptability and stability across market conditions are essential.

\begin{figure}
    \centering
    \includegraphics[width=1.00\linewidth]{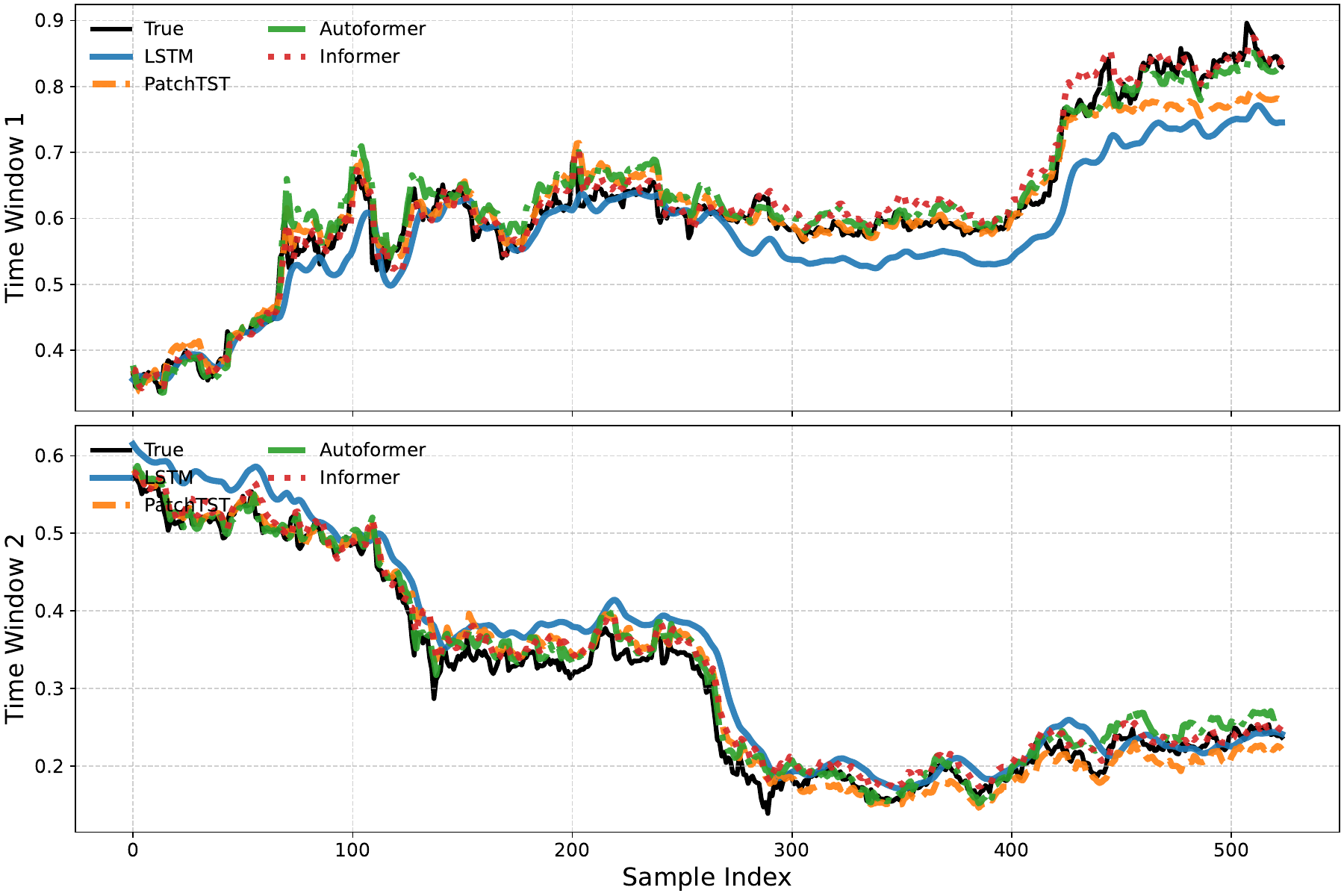}
    \caption{Comparison of \dkoopformer~on the test dataset for Crypto coin forecasting with $p = 40$, $H = 16$ and $d_{\text{model}}=48$, the LSTM
hidden size is $48$.}
    \label{fig:Crypto_coin_p_40_h_16_d_48}
\end{figure}

\begin{figure}
    \centering
    \includegraphics[width=1.00\linewidth]{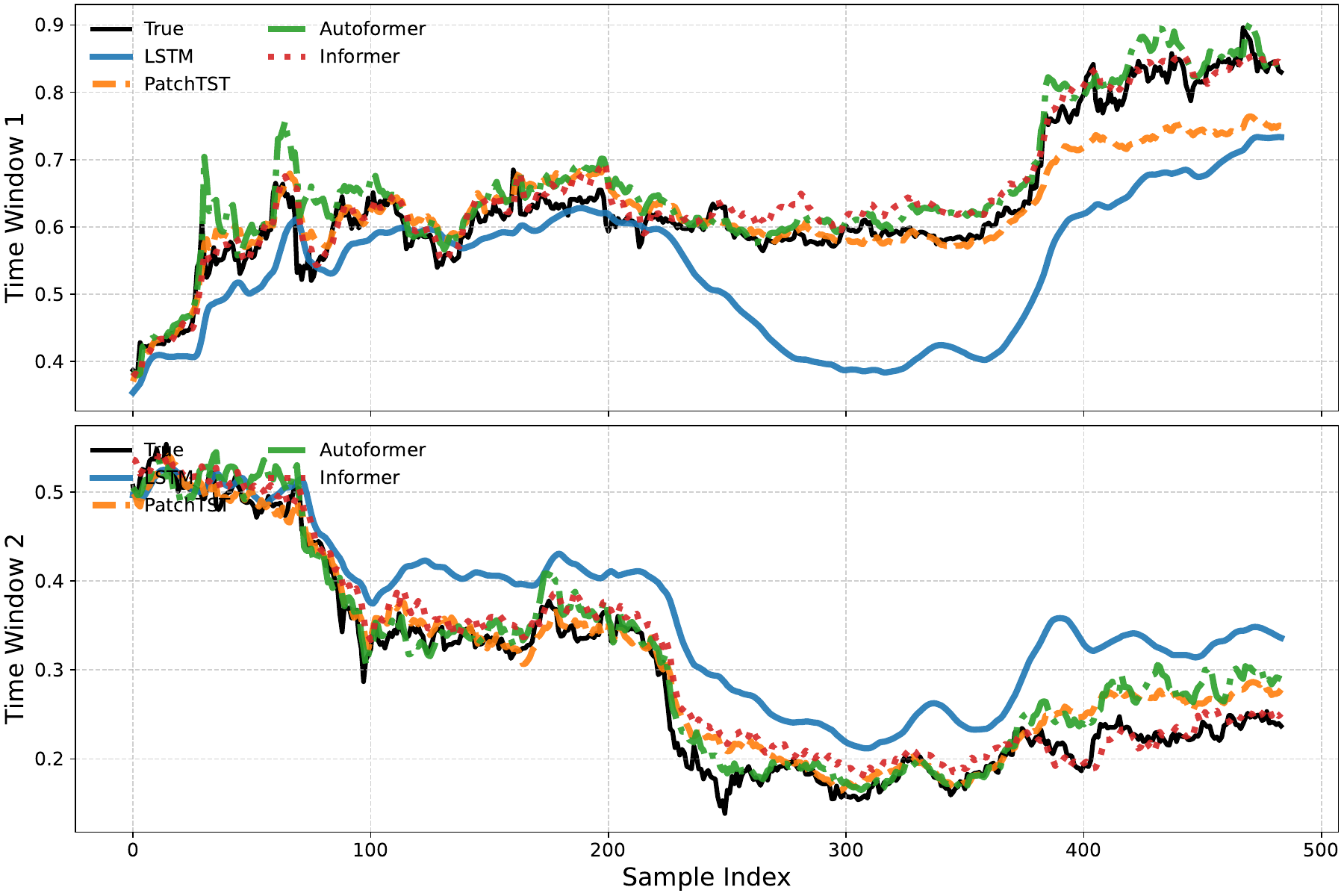}
    \caption{Comparison of \dkoopformer~on the test dataset for Crypto coin forecasting with $p = 80$, $H = 16$ and $d_{\text{model}}=48$, the LSTM
hidden size is $48$.}
    \label{fig:Crypto_coin_p_80_h_16_d_48}
\end{figure}

To further assess the forecasting accuracy of the competing models, we provide visual comparisons of their predictions against the ground truth in Figures~\ref{fig:Crypto_coin_p_40_h_16_d_48} and~\ref{fig:Crypto_coin_p_80_h_16_d_48}. These plots depict two representative time windows from the test set using a forecast horizon of $H = 16$ and latent dimension $d_{\text{model}} = 48$, with patch lengths $p = 40$ and $p = 80$, respectively.

Across both configurations, it is visually evident that the LSTM model consistently underperforms relative to the Transformer-based models. Specifically, LSTM exhibits a sluggish response to rapid trend reversals and fails to capture the amplitude of both upward and downward movements in the time series. This limitation is particularly apparent in Time Window 1 of Figure~\ref{fig:Crypto_coin_p_40_h_16_d_48}, where LSTM underestimates the series during a sharp decline. In contrast, \dkoopformer~variants, demonstrate closer alignment with the true signal, effectively tracking inflection points and local fluctuations.

The superior performance of \dkoopformer-based architectures is especially noticeable as the patch length increases (Figure~\ref{fig:Crypto_coin_p_80_h_16_d_48}), reinforcing their capacity to exploit richer temporal dependencies. Overall, these visual results corroborate the quantitative findings from the error heatmaps and highlight the limited generalization capability of LSTM in high-volatility, long-horizon financial forecasting tasks.

\subsection{Application of \dkoopformer~on electricity generation dataset}

In this scenario, we apply the \dkoopformer~model to the task of forecasting electricity generation from a variety of energy sources in Spain \footnote{\url{https://github.com/afshinfaramarzi/Energy-Demand-electricity-price-Forecasting/tree/main}}, including \textit{fossil fuels, wind, solar, hydro, and biomass}. The underlying dataset combines high-resolution hourly records of energy generation with corresponding weather data from multiple cities, forming a rich multivariate time series.

\subsubsection{Scenario 9}

We curated a subset of the original energy dataset across the recorded time horizon. The cleaned dataset focuses on electricity generation data from active sources, including \textit{biomass, fossil brown coal/lignite, fossil gas, fossil hard coal, fossil oil}, and \textit{geothermal energy}. Each of these generation types contributes a distinct temporal signature shaped by both dispatch policies and external conditions. The dataset comprises hourly measurements, and we selected a time window from index 4000 to 7500 to form a temporally consistent training segment that captures daily and weekly dynamics. 

This resulted in a five-channel array of shape $(3500,\, 6)$, which enabled the \dkoopformer~model to learn from parallel streams of electricity generation in a multi-channel forecasting setup. Leveraging this dataset, we train and evaluate the \dkoopformer~algorithm. The dataset is partitioned into training and testing splits with an 80\%/20\% ratio.


The \dkoopformer-based forecasting experiments explored a range of temporal and architectural settings to evaluate the models' generalization across different scales. In particular, the patch length—which determines the granularity of temporal segmentation or smoothing depending on the \dkoopformer~variant—was swept over the set $p= \{70, 80, 90, 100, 110, 120, 130\}$. This controls how much historical information is aggregated per token in \patchtst-style embeddings or how large the moving average window is in Autoformer-style decomposition. The forecast horizon is varied across $H=\{2, 4, 6, 8, 10, 12, 14, 16\}$, enabling performance evaluation on both short-term and extended look-ahead windows. All \dkoopformer~variants employed a latent dimension of \(d_{\text{model}} = 96\), which defines the embedding size used for temporal encoding, Koopman operator learning, and downstream prediction. These settings allow the models to balance expressive capacity with computational tractability across the tested configurations.

\begin{figure*}
    \centering
    \includegraphics[width=0.75\linewidth]{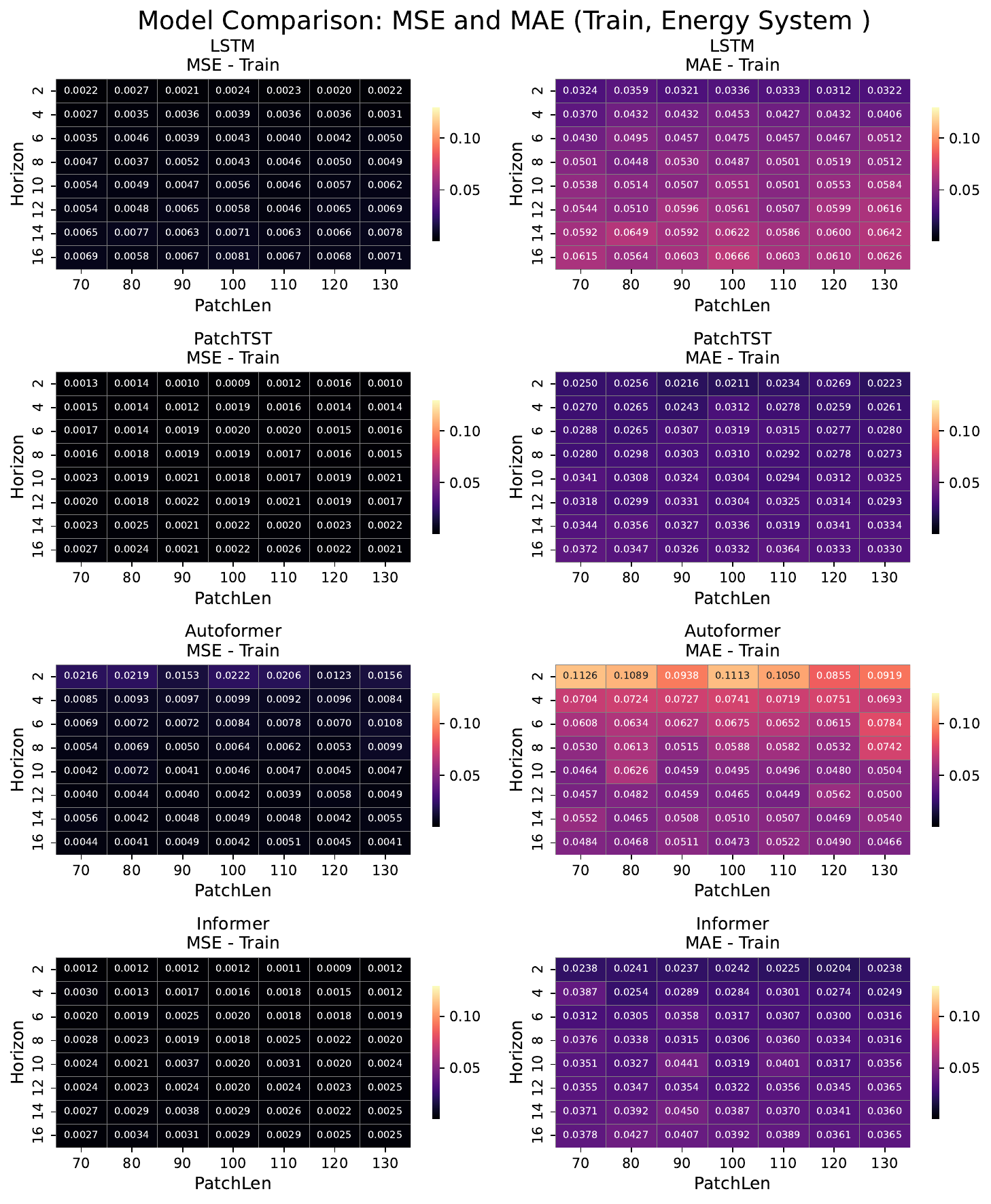}
    \caption{\dkoopformer~models versus LSTM baseline grid search in train dataset for electricity generation time series forecasting in scenario 9.}
    \label{fig:Energy_pred_train_grid}
\end{figure*}

Figure~\ref{fig:Energy_pred_train_grid} presents the training performance of different \dkoopformer-based forecasting models across various patch lengths and forecast horizons.

The training results reveal that \patchtst~achieves the lowest MSE and MAE across nearly all patch lengths and forecast horizons, highlighting its strong ability to capture temporal dependencies via patch-wise embeddings. \informer~shows competitive performance with moderate errors, particularly at shorter horizons, while \autoformer records higher MSE and MAE values overall, especially for longer horizons. This suggests that \dkoopformer-\patchtst~provides the most stable and accurate training convergence, followed by \informer, whereas \autoformer~is more sensitive to forecasting window and smoothing kernel size.
Moreover, \patchtst~and \informer outperforms the LSTM in almost all the settings, however that is not the case for \autoformer~specially for the shorter horizon, e.g. $H=2,\ 4$.

\begin{figure*}
    \centering
    \includegraphics[width=0.75\linewidth]{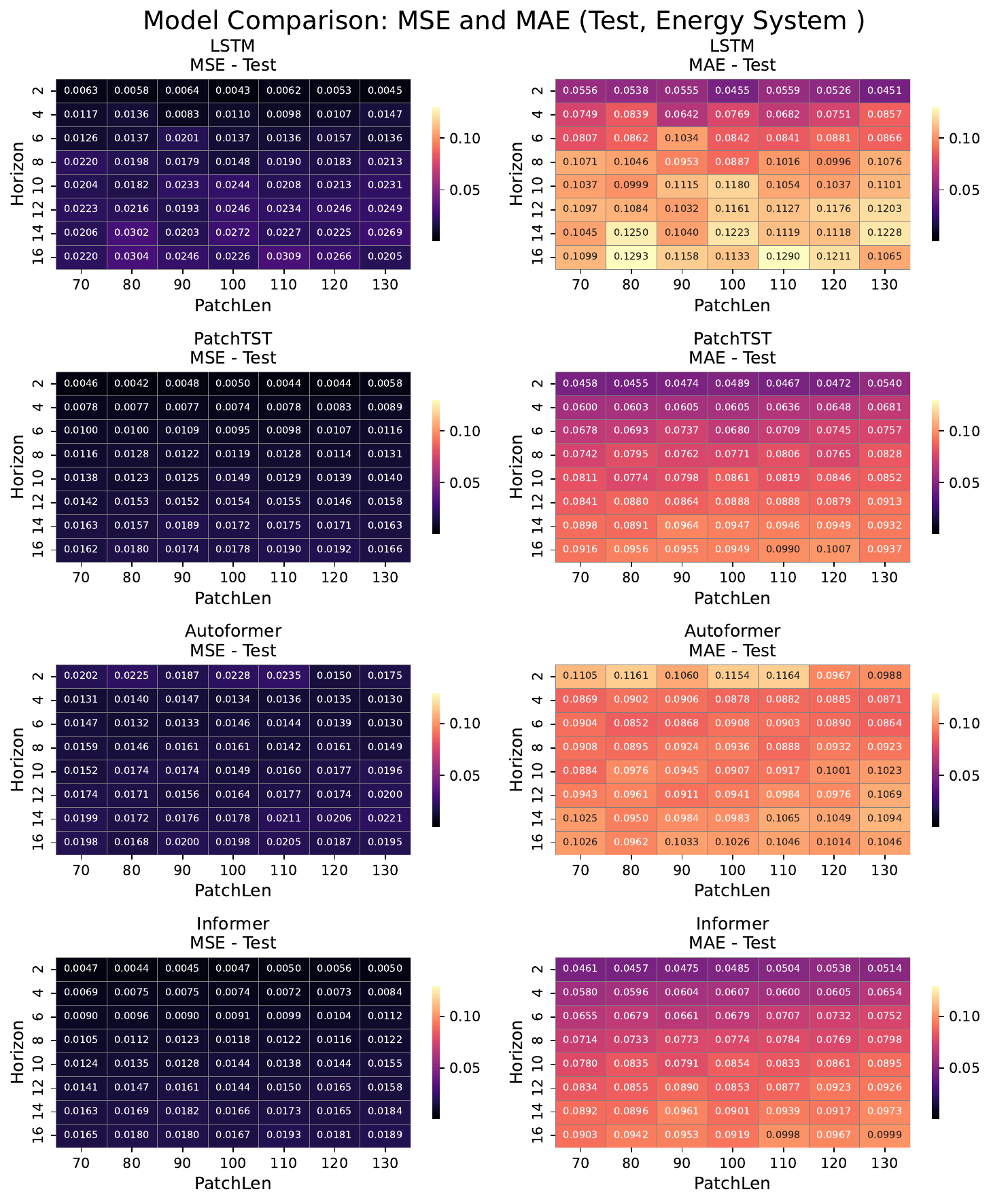}
    \caption{\dkoopformer~models versus LSTM baseline grid search in test dataset for electricity generation time series forecasting in scenario 9.}
    \label{fig:Energy_pred_test_grid}
\end{figure*}

As shown in Fig. \ref{fig:Energy_pred_test_grid} on the test set, \patchtst~again demonstrates superior generalization, achieving the lowest MSE and MAE across a wide range of patch lengths and forecast horizons. The error values remain stable even for long forecasting windows, indicating robust temporal representation and Koopman-stable dynamics. \informer~performs moderately well, especially at mid-range horizons, though its performance degrades slightly for long horizons. \autoformer, exhibits the highest test errors overall—particularly at larger patch sizes—suggesting over-smoothing in its trend extraction. Overall, \patchtst~yields the best balance between training convergence and test generalization, confirming its suitability for real-world energy forecasting tasks. The LSTM has the worst performance among all the models in the test dateset except for for short horizons, e.g. $H=2$, which has a competitive performance.

\begin{figure}
    \centering
    \includegraphics[width=0.99\linewidth]{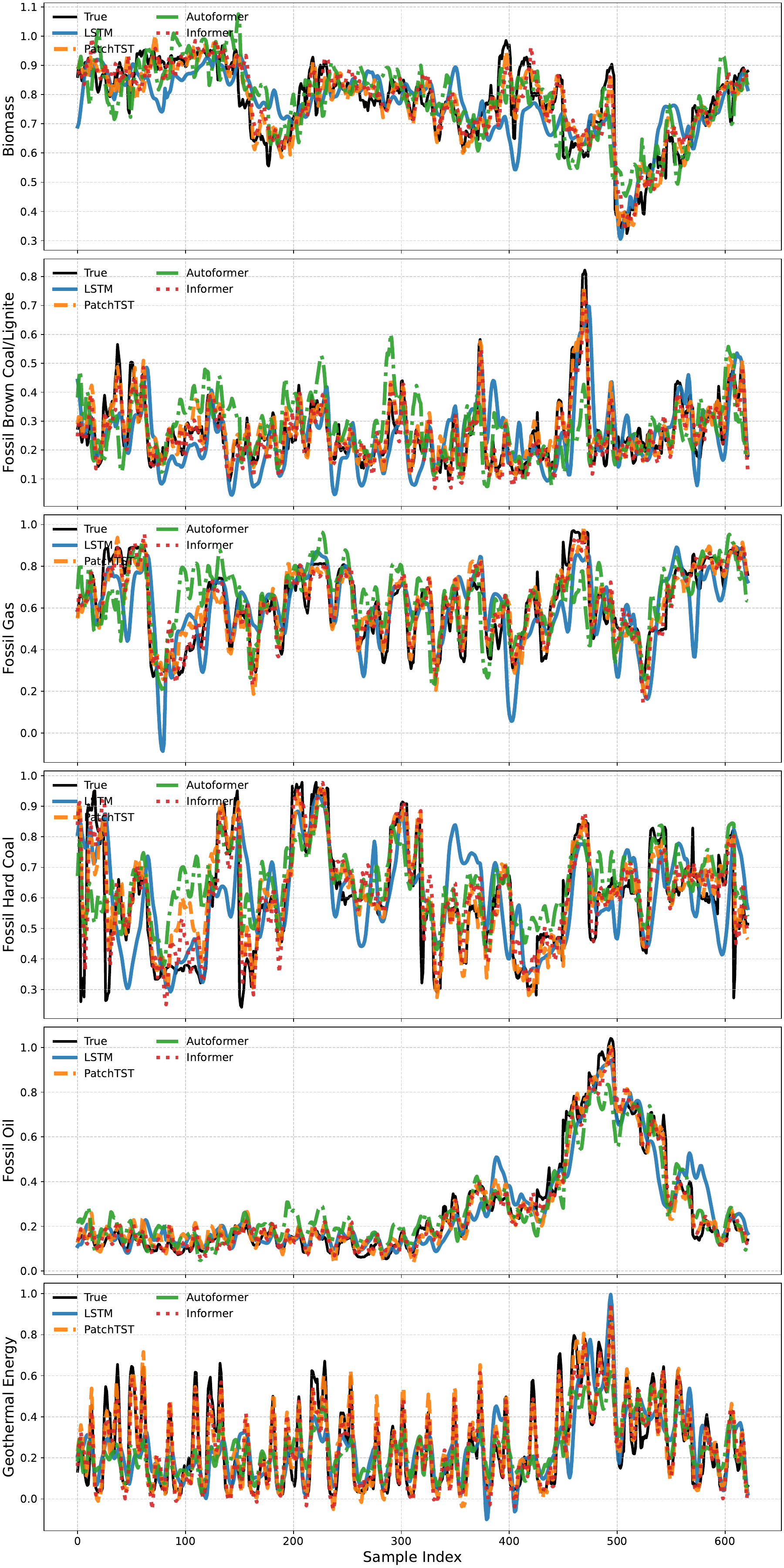}
    \caption{Comparison of \dkoopformer~on the test dataset for electricity generation forecasting with $p = 70$, $H = 8$ and $d_{\text{model}}=96$, the LSTM
hidden size is $96$.}
    \label{fig:forecast_comparison_energy_source_single_setting}
\end{figure}

Figure~\ref{fig:forecast_comparison_energy_source_single_setting} illustrates the forecasting performance of \dkoopformer~variants and the LSTM baseline on the test dataset for six different electricity generation sources, using a patch length $p=70$, forecast horizon $H=8$, and embedding dimension $d_{\text{model}}=96$. The \patchtst~shows the best alignment with the ground truth across all generation types, especially for highly dynamic signals like \textit{Fossil Gas}, \textit{Fossil Hard Coal}, and \textit{Geothermal Energy}. The \autoformer~tends to over-smooth rapid fluctuations, while the \informer~and LSTM follow trends moderately well but with reduced temporal accuracy. This visual comparison confirms the superior short-term predictive fidelity of \patchtst~and\informer~in modeling diverse power generation dynamics.

\section{Conclusion}\label{conclusion_sec}
This paper introduced \dkoopformer, a unified forecasting architecture that integrates Koopman operator theory with Transformer-based sequence modeling to address key limitations of current deep learning methods for time series forecasting. By enforcing spectral stability and energy dissipation through orthogonally parameterized Koopman operators and Lyapunov-inspired penalties, \dkoopformer~ensured both robust training and reliable multi-step predictions. Our contributions span theoretical, algorithmic, and empirical dimensions: we provided the first closed-form robustness guarantees for Transformer forecasters, decouple model capacity from latent stability via modular design, and release an extensible benchmark suite for Koopman-augmented variants of popular architectures such as \patchtst, \autoformer, and \informer.

Empirical results on nonlinear dynamical systems (Van der Pol and Lorenz) demonstrated the framework’s ability to learn structured latent dynamics even under noise. Large-scale experiments on CMIP6 and ERA5 climate datasets, cryptocurrency time series, as well as electricity generation time series further confirm its superior accuracy and robustness compared to conventional models, particularly in high-dimensional and multivariate forecasting settings. Our findings highlighted the value of embedding dynamical priors in deep architectures and open avenues for principled, interpretable time series forecasting across scientific domains. Future work extend this approach to control tasks, irregular time series, and spatiotemporal graph structures, further bridging the gap between data-driven learning and dynamical systems theory.


\appendix

\bibliographystyle{ieeetr}

\bibliography{refs/tutorial_bibliography,refs/koopman_ref_tutorial,refs/ref_koopformer, refs/references_brunton, refs/my_ref}


\end{document}